\documentclass[sigconf, nonacm]{acmart}
\usepackage{mathtools}
\usepackage{graphicx}
\usepackage{hhline}
\usepackage{makecell}
\usepackage{multicol}
\usepackage{multirow}
\usepackage{amsfonts}
\usepackage{caption}
\usepackage{subcaption}
\usepackage{changepage}
\usepackage{amsthm}
\usepackage{algpseudocode}
\usepackage{xspace}
\usepackage{colortbl}
\usepackage{tabularx}
\usepackage{tikz}

\newcommand{\spara}[1]{\smallskip\noindent{\bf #1}}

\newcommand{\para}[1]{\noindent{\bf #1}}

\algnewcommand\Params{\item[\textbf{Params:}]}

\definecolor{cycle1}{RGB}{228, 26, 28}
\definecolor{cycle2}{RGB}{55, 126, 184}
\definecolor{cycle3}{RGB}{77, 175, 74}
\definecolor{cycle4}{RGB}{152, 78, 163}
\definecolor{cycle5}{RGB}{255, 127, 0}
\definecolor{cycle6}{RGB}{153, 153, 153}
\definecolor{cycle7}{RGB}{166, 86, 40}
\definecolor{cycle8}{RGB}{247, 129, 191}

\newcommand*{\bigO}{\mathcal{O}}

\newcommand{\reals}{\mathbb{R}}

\newcommand{\dm}[1]{#1}

\newcommand{\ourmethod}{GDR\xspace}
\newcommand{\ourmethodU}{$\ourmethod_{\;UMAP}$\xspace}
\newcommand{\ourmethodN}{$\ourmethod_{\;TSNE}$\xspace}

\author{Andrew Draganov}
\affiliation{%
  \institution{Aarhus University}
   \country{}
}
\email{draganovandrew@cs.au.dk}

\author{Tyrus Berry}
\affiliation{\institution{George Mason University}
\country{}
}
\email{tberry@gmu.edu}

\author{Jakob Rødsgaard Jørgensen}
\affiliation{%
  \institution{Aarhus University}
    \country{}
}
\email{jakobrj@cs.au.dk}

\author{Katrine Scheel Nellemann}
\affiliation{%
  \institution{Aarhus University}
    \country{}
}
\email{scheel@cs.au.dk}

\author{Ira Assent}
\affiliation{%
  \institution{Aarhus University}
    \country{}
}
\email{ira@cs.au.dk}

\author{Davide Mottin}
\affiliation{%
  \institution{Aarhus University}
    \country{}
}
\email{davide@cs.au.dk}

\author{Cigdem Aslay}
\affiliation{%
  \institution{Aarhus University}
    \country{}
}
\email{cigdem@cs.au.dk}

\begin{document}
\title{GiDR-DUN: Gradient Dimensionality Reduction-\\ Differences and Unification}

\begin{abstract}
TSNE and UMAP are two of the most popular dimensionality reduction algorithms due to their speed and interpretable low-dimensional embeddings. However, while attempts have been made to improve on TSNE’s
computational complexity, no existing method can obtain TSNE embeddings at the speed of
UMAP. In this work, we show that this is indeed possible by combining the two approaches into a single method. We theoretically and experimentally evaluate the full space of parameters in the TSNE and UMAP algorithms and observe that a single parameter -- the normalization -- is responsible for switching between them. This, in turn, implies that a majority of the algorithmic differences can be toggled without affecting the embeddings. We discuss the implications this has on several theoretic claims underpinning the UMAP framework, as well as how to reconcile them with existing TSNE interpretations.

Based on our analysis, we propose a new dimensionality reduction algorithm, \ourmethod, that combines previously incompatible techniques from TSNE and UMAP and can replicate the results of either algorithm by changing the normalization. As a further advantage, \ourmethod performs the optimization faster than available UMAP methods and thus an order of magnitude faster than available TSNE methods. Our implementation is plug-and-play with the traditional UMAP and TSNE libraries and can be found at \url{https://github.com/Andrew-Draganov/GiDR-DUN}.

\end{abstract}

\maketitle

\section{Introduction}
Dimensionality Reduction (DR) algorithms are invaluable for qualitatively inspecting high-dimensional data and are widely used across scientific disciplines. This includes tasks such as visualizing single-cell RNA-sequence data \cite{xiang2021comparison}, understanding molecular dynamics \cite{trozzi2021umap}, and classifying toxicological data \cite{lovric2021should}, to name a few. At a broad level, DR algorithms accept a high-dimensional input and find a faithful embedding in a lower-dimensional space. This embedding aims to preserve similarities among the $n$ points, where similarity is measured by distances in the corresponding spaces. This implies a quadratic run time, as DR algorithms must compare all $n^2$ high- and low-dimensional pairwise similarities before we can be certain of the embedding quality.

\newcolumntype{C}{ >{\centering\arraybackslash} m{3.25cm} }
\newcolumntype{D}{ >{\centering\arraybackslash} m{2cm} }
\begin{table*}[h]
    \begin{tabularx}{\textwidth}{DCCCC}
        \textit{dataset} & TSNE & \ourmethodN &  UMAP & \ourmethodU \\
        \midrule
        
        \makecell{MNIST} &
        \includegraphics[width=.77\linewidth]{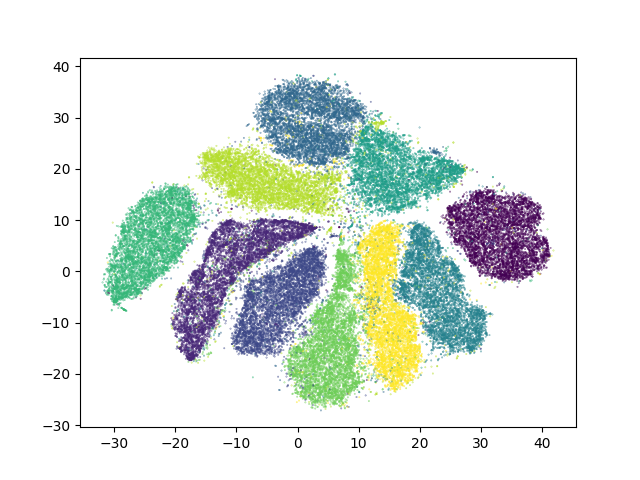} &
        \includegraphics[width=.77\linewidth]{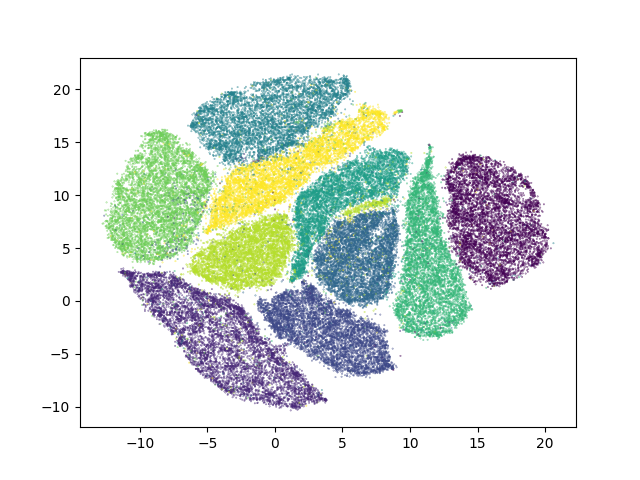} &
        \includegraphics[width=.77\linewidth]{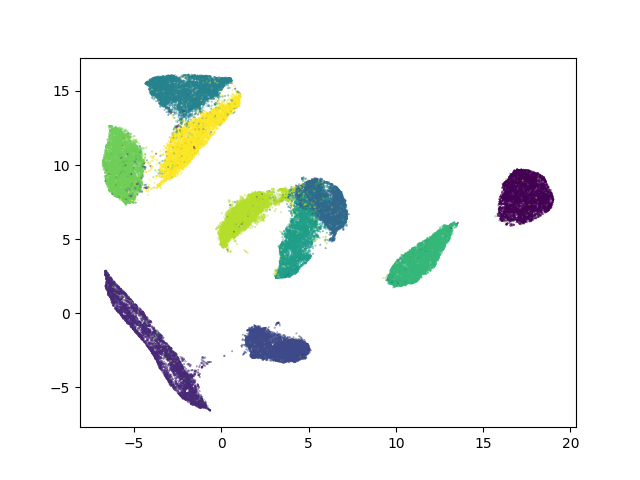} & 
        \includegraphics[width=.77\linewidth]{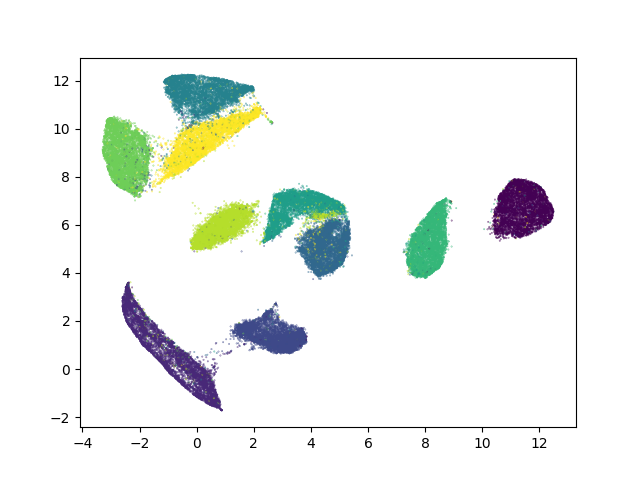} \\
        
        \makecell{Fashion MNIST} &
        \includegraphics[width=.77\linewidth]{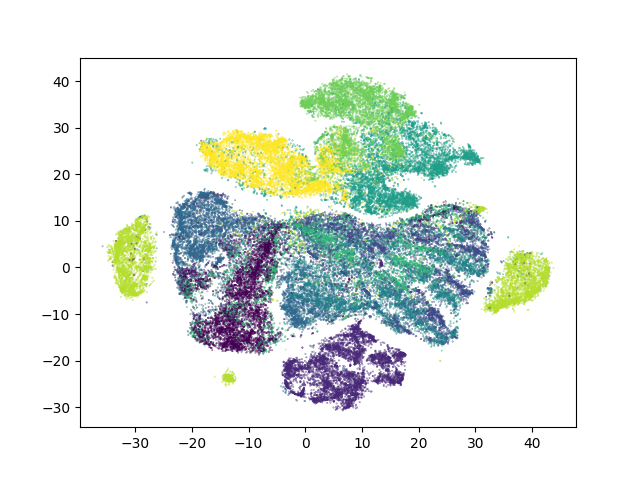} &
        \includegraphics[width=.77\linewidth]{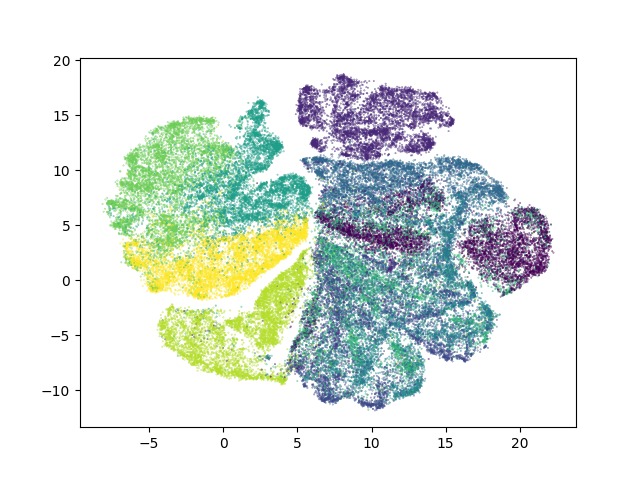} &
        \includegraphics[width=.77\linewidth]{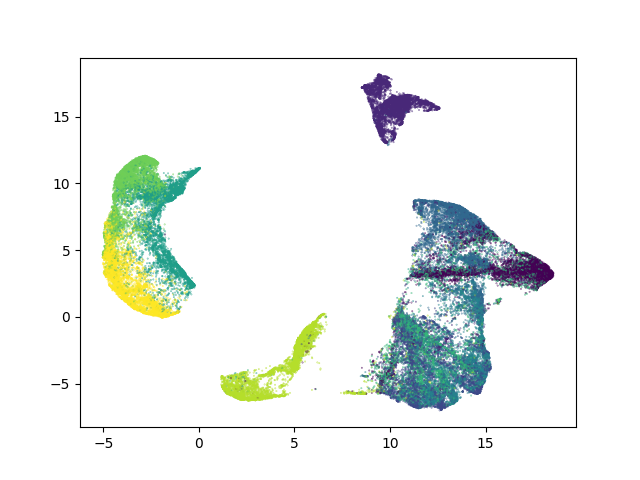} & 
        \includegraphics[width=.77\linewidth]{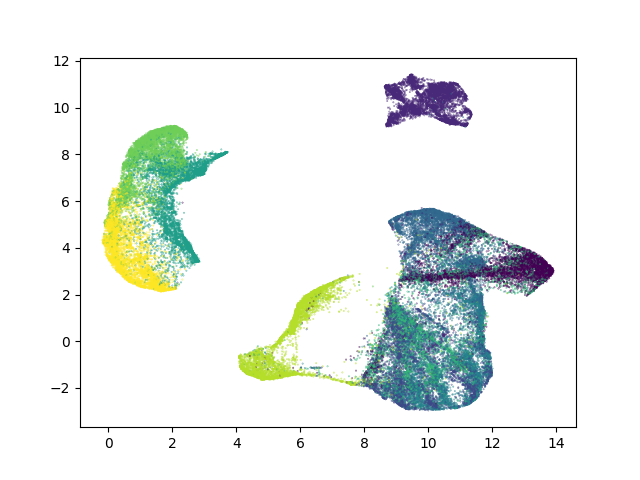} \\
        
        \makecell{Coil-100} &
        \includegraphics[width=.77\linewidth]{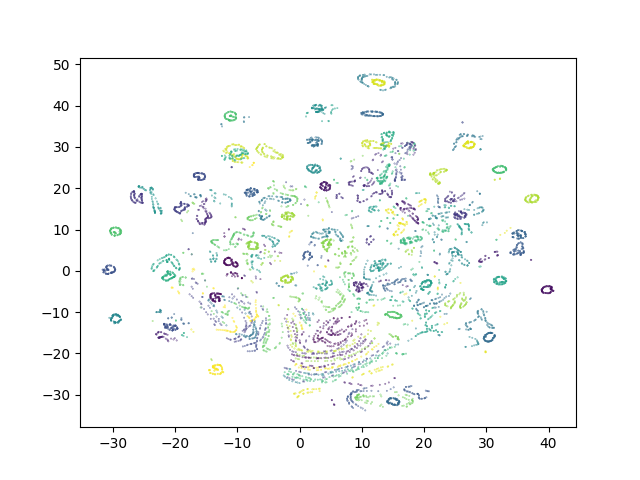} &
        \includegraphics[width=.77\linewidth]{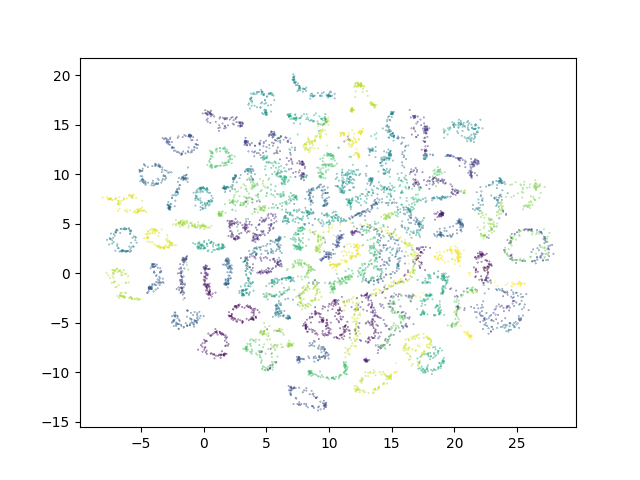} &
        \includegraphics[width=.77\linewidth]{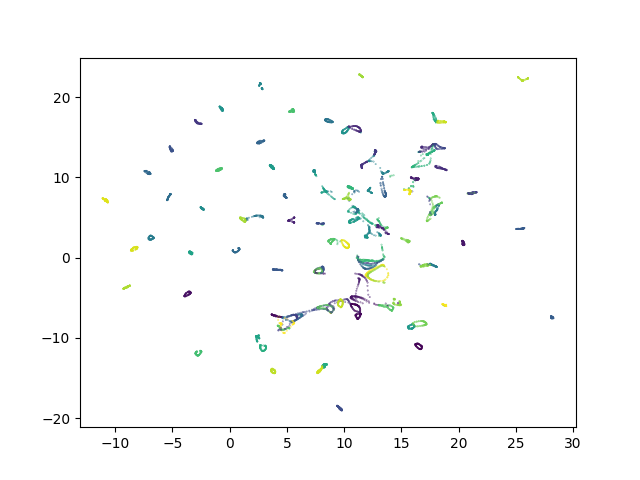} &
        \includegraphics[width=.77\linewidth]{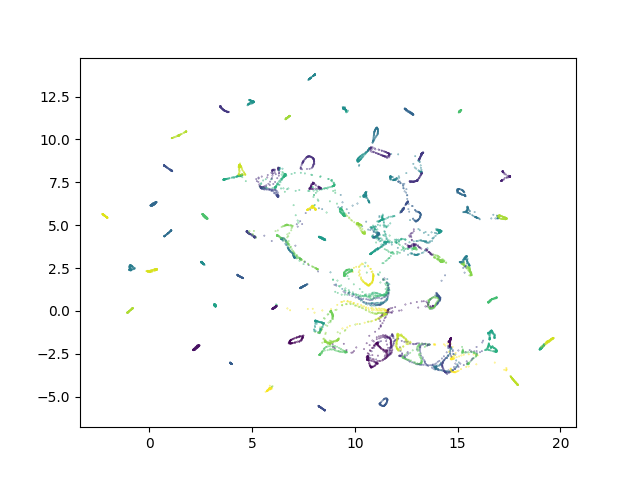} \\
        \bottomrule
    \end{tabularx}
\caption{\ourmethod recreates both TSNE and UMAP embeddings by changing the normalization and corresponding gradients.}
\end{table*}

Recent gradient-based methods have improved this run time by performing the optimization on only the relevant entries of the similarity matrix.
Among these, TSNE~\cite{van2008visualizing, van2014accelerating} and UMAP~\cite{mcinnes2018umap} are by far the most popular due to their improved speed and clusterable embeddings. Both algorithms operate similarly. They establish analogous similarity measures, find an embedding through gradient descent, and even share comparable loss functions. Despite these similarities, TSNE and UMAP have several key differences. First, although both methods obtain qualitatively similar results, UMAP prefers large inter-cluster distances while TSNE leans towards large intra-cluster distances. Second, UMAP runs significantly faster as it no longer needs to partition the low-dimensional space during every epoch of gradient descent like TSNE does. While attempts have been made to improve on TSNE's computational complexity \cite{linderman2019fast,tang2016visualizing}, there has not yet been a method that can obtain both TSNE and UMAP embeddings at the speed of UMAP.

We believe that this is in part due to their radically different presentations. While TSNE is introduced from a computational perspective, UMAP is instead introduced with a deep foundation in category theory and topology. Despite this, many of the algorithmic choices in UMAP and TSNE are presented without theoretical justification, making it difficult to know which components of each algorithm are necessary. For example, UMAP suggests that the KL-divergence is a natural objective function due to its underlying theory. However, it is unclear whether this is a necessary choice; indeed we show that one can minimize the Frobenius norm in both TSNE and UMAP and obtain equivalent embeddings. Such questions can also be raised for many other algorithmic components such as the normalization, symmetrizations, and initialization requirements, which we will discuss in later sections. 


In this paper, we make the argument that the differences in \textit{both} the embeddings \textit{and} computational complexity between TSNE and UMAP can be attributed to a single algorithmic choice -- the normalization factor. We come to this conclusion by first identifying every framework and implementation difference between the two methods, thus defining both algorithms as a set of on/off switches across the space of these parameters. Under this lens, we implement TSNE and UMAP in a common library, giving us the ability to study the effect that each such choice has on the embeddings. We show both quantitatively and qualitatively that the majority of these parameters have no distinguishable effect on the resulting embeddings, with the exception of the normalization of the pairwise similarity matrices $P$ and $Q$. This is substantiated through theoretical discussion, where we analyze the effect that the normalization has on the resulting gradient descent schema. Thus, we show that making each algorithm compatible with the other's normalization is a sufficient condition for UMAP to give TSNE embeddings and for TSNE to give UMAP embeddings.

This leads us to the conclusion that many of the differences between TSNE and UMAP are negligible and, in turn, brings several claims regarding the necessary components of each algorithm into question. Furthermore, we believe that UMAP's theoretic foundation does not depend on the normalization, raising questions on whether TSNE can be interpreted under UMAP's topological framework.

Based on this analysis we propose \ourmethod, a DR algorithm that can perform either TSNE or UMAP depending on the choice of normalization. Since the normalization is a sufficient condition for switching between the algorithms, we are then free to make computationally efficient choices among the remaining hyperparameters. Due to this, \ourmethod is particularly amenable to parallelization, running at UMAP speed despite making fewer estimates in many of the gradient calculations. We experimentally validate that \ourmethod can simulate both methods through a thorough quantitative and qualitative evaluation. Furthermore, our method is faster than any available version of UMAP and TSNE.

In summary, our contributions are as follows:
\begin{enumerate}
        \item We perform the first broad analysis of the conceptual and algorithmic differences between TSNE and UMAP, showing the effect of each algorithmic choice on the embeddings.
        \item We propose \ourmethod, an optimization algorithm that effectively reproduces both TSNE and UMAP embeddings depending on a single input parameter.
        \item We show that \ourmethod is faster than available UMAP methods and thus significantly faster than all TSNE methods.
        \item We release simple, plug-and-play implementations of \ourmethod, TSNE and UMAP that can each toggle all of the hyperparameters we've identified.
\end{enumerate}

\section{Related Work}

\ourmethod falls into a broad category of gradient-based DR approaches, characterized by minimizing a non-convex objective using variants of gradient descent.
The basic steps in TSNE and UMAP consist of finding nearest neighbors in the high-dimensional space, obtaining similarity matrices $P$ and $Q$ between pairs of points in high- and low-dimensions, and optimizing $KL(P||Q)$ with gradient descent. These steps are discussed in more detail in Section~\ref{comparison}.

There exist several variations to these methods. We first mention that when discussing TSNE, we
are referring to~\cite{van2014accelerating}, which established the nearest neighbor and sampling speed improvements. Since this paper, a popular subsequent development
was presented in~\cite{linderman2019fast}, wherein Fast Fourier Transforms were used to accelerate the comparisons between points. Another approach based on
TSNE is the LargeVis algorithm, proposed in ~\cite{tang2016visualizing}, which modifies the embedding functions to satisfy a graph-based Bernoulli probabilistic model of
the low-dimensional dataset. As the more recent algorithm, UMAP has not had as many variations yet. One promising direction, however, has been to extend the
second step of UMAP as a parametric optimization on neural network weights~\cite{sainburg2020parametric}.

Many of these approaches utilize the same optimization structure where they iteratively attract and repel points. While most perform their attractions similarly
along nearest neighbors in the high-dimensional space, the repulsions are the slowest operation and each method approaches the repulsions differently.
The TSNE method we are discussing samples repulsions by utilizing Barnes-Hut trees to sum their effects of distant points.
The work in~\cite{linderman2019fast} instead calculates repulsive forces with respect to specifically chosen interpolation points, cutting down on the $\bigO(n\log n)$
requirement of the tree calculations at every gradient step. UMAP and LargeVis, on the other hand, simplify the repulsion sampling by only calculating the
gradient with respect to a constant number of points. These repulsion techniques are, on their face, incompatible with one another, i.e., several modifications have to be made
to each algorithm before one can interchange the repulsive force calculations.

While modifying the algorithms is more challenging, there have been several papers that attempt to explain the difference between TSNE and UMAP by looking at
other design choices. The authors in both~\cite{kobak2019umap} and~\cite{kobak2021initialization},
make the argument that the necessary-and-sufficient condition for switching between their embeddings is the difference in initialization.
Namely, TSNE randomly initializes the low dimensional embedding whereas UMAP calculates a Laplacian Eigenmap~\cite{belkin2003laplacian} embedding before
beginning the optimization. We have been unable to reproduce these results across datasets and optimization criteria, a topic we discuss later in our results.

Several other works have compared the UMAP and TSNE embeddings through a more theoretical analysis~\cite{damrich2021umap, bohm2020unifying}. The authors in  \cite{damrich2021umap} find that UMAP's true optimization calculates a different value than is proposed in the original paper. This is in line with our findings, where we show that UMAP only converges when performing a constant number of gradient calculations per epoch despite requiring an $\bigO(n)$ number of repulsions according to its derivations. A similar argument is made in \cite{bohm2020unifying}, where TSNE and UMAP are analyzed through their attractive and repulsive forces. We expand on the key finding developed there, that the ratio between attractions and repulsions is fundamental to the structure of the embedding. Indeed, we show that these repulsions and attractions are solely determined by the choice of normalization, thus giving a practical treatment of their proposed ideas.

Lastly, we also mention the state-of-the-art implementations of both TSNE and UMAP. We implement the Barnes-Hut variant of TSNE and the traditional version of UMAP in order to do controlled comparisons. We note that our implementations are faster than both the traditional \texttt{scikit-learn} and \texttt{umap-learn} implementations. We also note that there exist several implementations specifically designed for the computing model of the GPU which can be found in the RAPIDS AI framework \cite{nolet2021bringing,rapidsframework}.
\section{Comparison of TSNE and UMAP} \label{comparison}

We begin by formally introducing the TSNE and UMAP DR algorithms. Let $X \in \reals^{n \times D}$ be a high dimensional dataset of $n$ points and let $Y\in \reals^{n\times d}$ be a previously initialized set of $n$ points in lower-dimensional space such that $d \ll D$. Our aim is to first define similarity measures between the points in each space and then find the embedding $Y$ such that the pairwise similarities in $Y$ match those in $X$.

To do this, both algorithms define high- and low-dimensional non-linear functions $p: X \times X \rightarrow [0, 1]$ and $q: Y \times Y \rightarrow [0, 1]$, forming pairwise similarity matrices $P(X)$ and $Q(Y)$, both in $\reals^{n \times n}$. Each element $p_{ij} \in P$ represents similarity between $x_i$ and $x_j$, while each $q_{ij} \in Q$ represents similarity between $y_i$ and $y_j$. Formally, define $p$ and $q$ as
\begin{equation}
\begin{aligned}
    p^{tsne}_{j|i}(x_i, x_j) &= \dfrac{\text{exp}(-d(x_i, x_j)^2 / 2 \sigma_i^2)}{\sum_{k \neq l} \text{exp}(-d(x_k, x_l)^2 / 2 \sigma_k^2)} \\[0.5ex]
    q^{tsne}_{ij}(y_i, y_j) &= \dfrac{(1 + ||y_i - y_j||^2_2)^{-1}}{\sum_{k \neq l} (1 + ||y_k - y_l||^2_2)^{-1}}
\end{aligned}
\label{eq:tsne_prob}
\end{equation}
\begin{equation}
\begin{aligned}
    p^{umap}_{j|i}(x_i, x_j) &= \text{exp} \left( (-d(x_i, x_j)^2 + \rho_{i}) /\tau_i \right) \\[0.3ex]
    q^{umap}_{ij}(y_i, y_j) &= \left( 1 + a(||y_i - y_j||^2_2)^b \right) ^{-1},
\end{aligned}
\end{equation}
where $d(x_i, x_j)$ is the high-dimensional distance function, $\sigma$ and $\tau$ are point-specific variance scalars, $\rho_i = \min_{j \neq i} d(x_i, x_j)$, and $a$ and $b$ are constants. Note that the denominators in the TSNE functions in Equation~\ref{eq:tsne_prob} are  $\sum P$ and $\sum Q$. We thus refer to TSNE's similarity functions as being \emph{normalized} while UMAP is \emph{unnormalized}.

In practice, we can assume that $2 \sigma_i^2$ is functionally equivalent to $\tau_i$, as they are both chosen such that the entropy of the resulting distribution is equivalent. The high-dimensional $p$ values are defined with respect to the point in question but are symmetrized by applying symmetrization functions. Without loss of generality, let $p_{ij}
= S(p_{j|i}, p_{i|j})$ for some symmetrization function $S$. Going forward, we write $p_{ij}$ and $q_{ij}$ without the superscripts when discussing them as the general high- and low-dimensional similarity functions.

Given these pairwise similarities in the high- and low-dimensional spaces, TSNE and UMAP attempt to find the embedding $Y$ such that $Q(Y)$ is closest to $P(X)$ under some cost function. Both similarity matrices, up to normalization, carry a probabilistic interpretation, allowing both TSNE and UMAP to model the dimensionality reduction problem as minimizing the KL divergence $KL(P \| Q)$. This gives us the following loss functions:
\begin{align}
    \mathcal{L}_{tsne} &= \sum_{i \neq j} p_{ij} \log \dfrac{p_{ij}}{q_{ij}} \\
    \mathcal{L}_{umap} &= \sum_{i \neq j} \left[ p_{ij} \log \dfrac{p_{ij}}{q_{ij}} + (1 - p_{ij}) \log \dfrac{1 - p_{ij}}{1 - q_{ij}} \right] 
\end{align}
In essence, TSNE minimizes the KL divergence of the entire pairwise similarity matrix since both $P$ and $Q$ sum to $1$ under TSNE. UMAP instead defines Bernoulli probability distributions $\{p_{ij}, 1-p_{ij}\}, \{q_{ij}, 1 - q_{ij}\}$ and sums the KL divergences between the $n^2$ pairwise probability distributions \footnote{We note that both TSNE and UMAP set the diagonals of $P$ and $Q$ to $0$, as they would otherwise trivially equal $1$}. Under these interpretations, we can think of $P$ and $Q$ as defining a graph over the points $X$ and $Y$, with edges $e_{ij}$ being equal to the similarity values $p_{ij}$ or $q_{ij}$. The probabilities in TSNE's normalized case are then akin to asking ``when we sample a random edge, what is the probability that we pick $e_{ij}$?''. Alternatively, the unnormalized UMAP variant asks ``for the specific pair of points $i$ and $j$, what is the probability that the edge $e_{ij}$ exists?''. As discussed in \cite{van2014accelerating}, we can interpret the gradient descent problem as a set of springs along each edge $e_{ij}$, with the spring constant being determined as a function of their high- and low-dimensional similarities $p_{ij}$ and $q_{ij}$.

\begin{figure}[htb]
    \includegraphics[width=.95\linewidth]{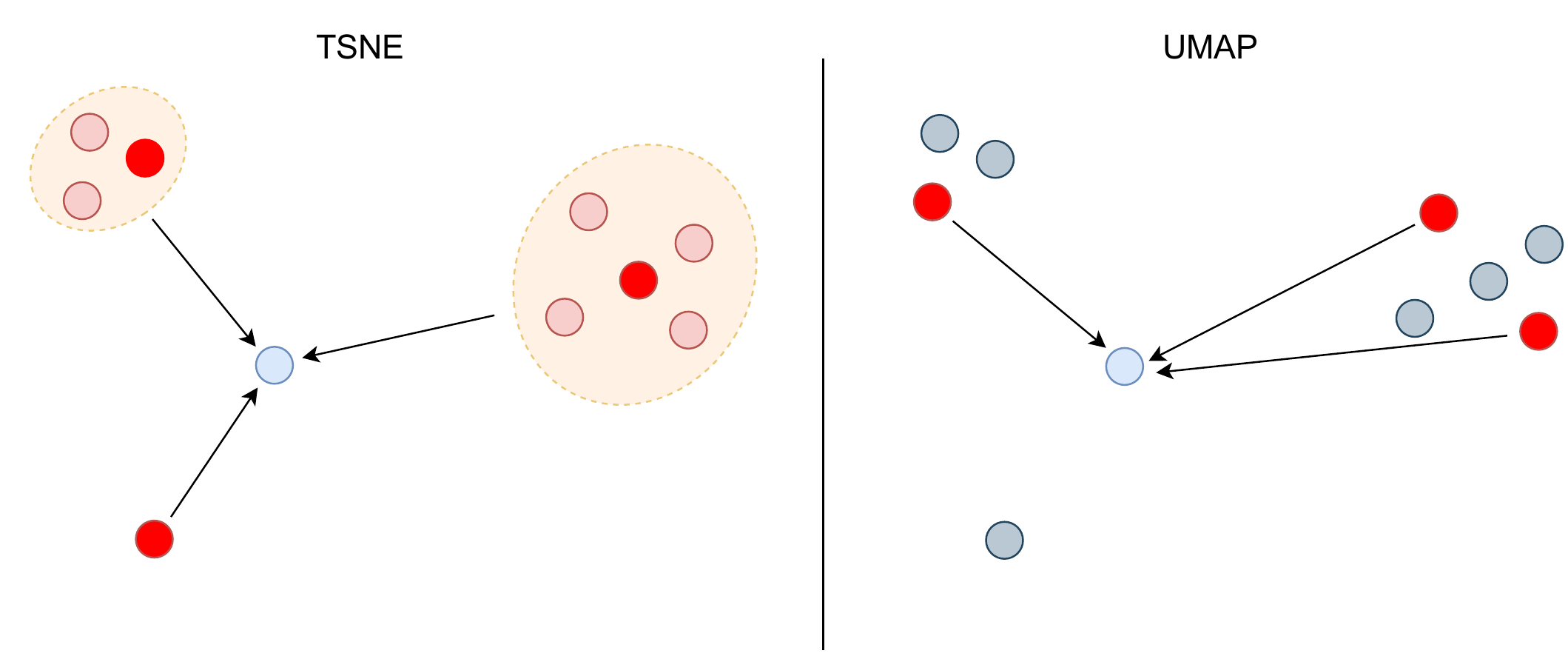}
    \caption{Visualization of the repulsive forces in TSNE (left) and UMAP (right). In TSNE, we calculate the repulsion for representative points and use this as a proxy for nearby points, giving $\bigO(n)$ total repulsions acting on each point. In UMAP, we calculate the repulsion to a pre-defined number of points and ignore the others, giving $\bigO(1)$ per-point repulsions. Bright red points are those for which the gradient is calculated; arrows are the direction of repulsion.}
    \label{repulsion_vis}
\end{figure}

\subsection{Gradient Calculations}
\label{grad_calc_sec}

We now describe and analyze the gradient descent approaches in TSNE and UMAP in order to identify why their embeddings have different intra- vs. inter-cluster distances. First, notice that the gradients of each algorithm change substantially due to the differing normalizations. In TSNE, the gradient can be written as
\begin{equation}
    \dfrac{\partial \mathcal{L}_{tsne}}{\partial y_i} = -4 \sum_{j \neq i} (p_{ij} - q_{ij}) q_{ij} Z (y_i - y_j),
    \label{tsne_grad_equations}
\end{equation}
where $Z = \sum_{k \neq l} (1 + ||y_k - y_l||_2^2)^{-1}$ is the normalization factor for the low-dimensional kernel. This is often represented as an attractive
and repulsive force with
\begin{align*}
    \dfrac{\partial \mathcal{L}_{tsne}}{\partial y_i} &= -4(\mathcal{A}_i^{tsne} + \mathcal{R}_i^{tsne}) = \\
    &= -4 \left[ \sum_{j, j \neq i} p_{ij}q_{ij}Z (y_i - y_j) - \sum_{k, k \neq i} q_{ik}^2 Z (y_i - y_k) \right.]
\end{align*}

UMAP also describes, separating its gradient into attractive and repulsive terms, as
\begin{align}
    \mathcal{A}_i^{umap} = & \sum_{j, j \neq i} \dfrac{-2ab\|y_i - y_j\|_2^{2(b-1)}}{1 + \|y_i - y_j\|_2^2} p_{ij} (y_i - y_j) \label{umap_attr} \\
    \mathcal{R}_i^{umap} = & \sum_{k, k \neq i} \dfrac{2b}{\epsilon + \|y_i - y_k\|_2^2} q_{ik} (1 - p_{ik}) (y_i - y_k). \label{umap_rep}
\end{align}

In practice, TSNE and UMAP optimize their loss functions by iteratively applying these attractive and repulsive forces.
It is unnecessary to calculate each such force to effectively estimate the gradient, however,
as the $p(x_i, x_j)$ multiplicative factor in both the TSNE and UMAP attractive forces decays exponentially.
Based on this observation, both approaches establish a nearest neighbor graph \cite{van2014accelerating} $G = (X, \mathcal{E})$ in the high-dimensional space, where edges represent nearest
neighbor relationships from $x_i$ to $x_j$ and vice versa. It then suffices to only perform attractions between points $y_i$ and $y_j$ if their corresponding $x_i$ and $x_j$ are nearest neighbors.

This sampling scheme does not transfer to the repulsions, however, as the Student-t distribution has a heavier tail. This means that repulsions must be calculated evenly across the rest of the points. TSNE does this by calculating a Barnes-Hut tree across $Y$ for every epoch. If $y_j$ and $y_k$ are both in the same cell then we assume $q_{ij} = q_{ik}$, allowing us to only calculate $\sim \log(n)$ similarities and apply them across the rest of the points. TSNE thus estimates all $n-1$ repulsions by performing one such estimate for each cell in $Y$'s Barnes-Hut tree. UMAP, on the other hand, simply obtains repulsions by sampling points uniformly and only applying those repulsions. These repulsion schemas are depicted in Figure~\ref{repulsion_vis}.

Given these attractive and repulsive forces, TSNE and UMAP proceed by performing gradient descent on the points. TSNE does this by collecting all of the gradients and performing momentum gradient descent across the entire dataset whereas UMAP moves each point immediately upon calculating its forces. Beyond this, there are multiple other differences in their gradient descent strategies. First, the TSNE learning rate stays constant over the course of training while UMAP linearly decreases the learning rate. Second, TSNE's gradients are further strengthened by adding a ''gains`` term which scales gradients based on whether they point in the same direction from epoch to epoch\footnote{This term has not been mentioned in the literature but is present across common TSNE implementations}. We refer to TSNE's gradient descent methodology as \textit{gradient amplification}.

We draw the reader's attention to the fact that the $j$ and $k$ iterands represent different samplings. Specifically, $j$ represents $i$'s nearest neighbor whereas $k$ is a random point with respect to $i$. We mention this to emphasize that the $p_{ij}$'s in UMAP's attractive force have been precomputed during the nearest neighbor graph construction whereas the corresponding $p_{ik}$'s are unknown during the gradient descent process. In practice, UMAP estimates the $p_{ik}$ values by substituting the available $p_{ij}$'s in instead. We also note that in practice UMAP does not explicitly multiply by $p_{ij}$ and $1 - p_{ik}$. Instead, it samples the attractions and repulsions proportionally to these multiplicative scalars. We refer to this as \textit{scalar sampling}. TSNE, however, calculates its gradient by performing the relevant multiplications.


\begin{figure*}[h]
    \newcolumntype{C}{ >{\centering\arraybackslash} m{3.35cm} }
    \newcolumntype{D}{ >{\centering\arraybackslash} m{2cm} }

    \begin{tabular}{DCCCC}
        & \multicolumn{2}{c}{\textit{Linearly growing distances}} & \multicolumn{2}{c}{\textit{Exponentially growing distances}} \\
        \cmidrule(lr){2-3}\cmidrule(lr){4-5}
        & TSNE & UMAP & TSNE & UMAP \\
        KL-Divergence &
        \includegraphics[width=\linewidth]{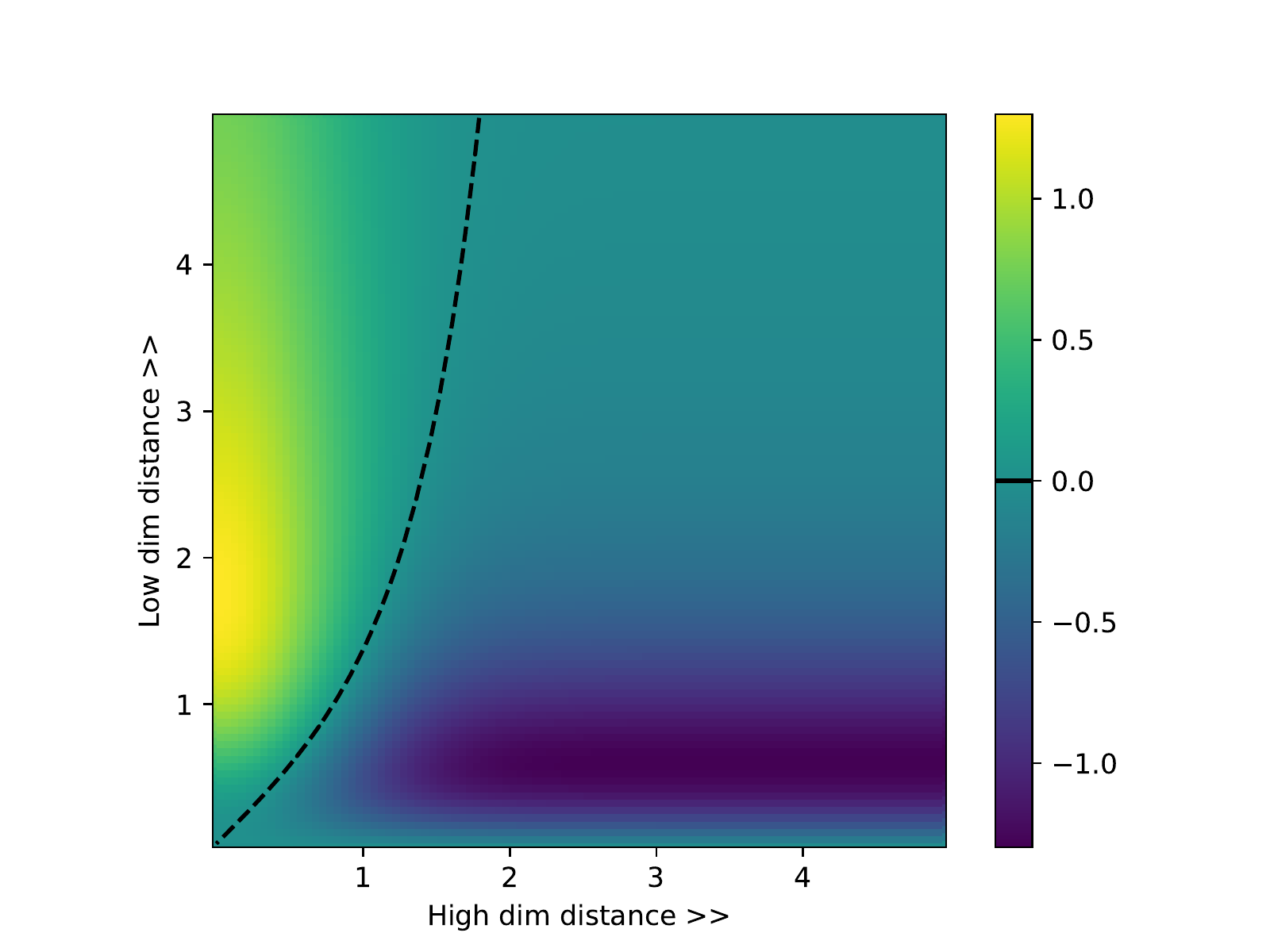} & 
        \includegraphics[width=\linewidth]{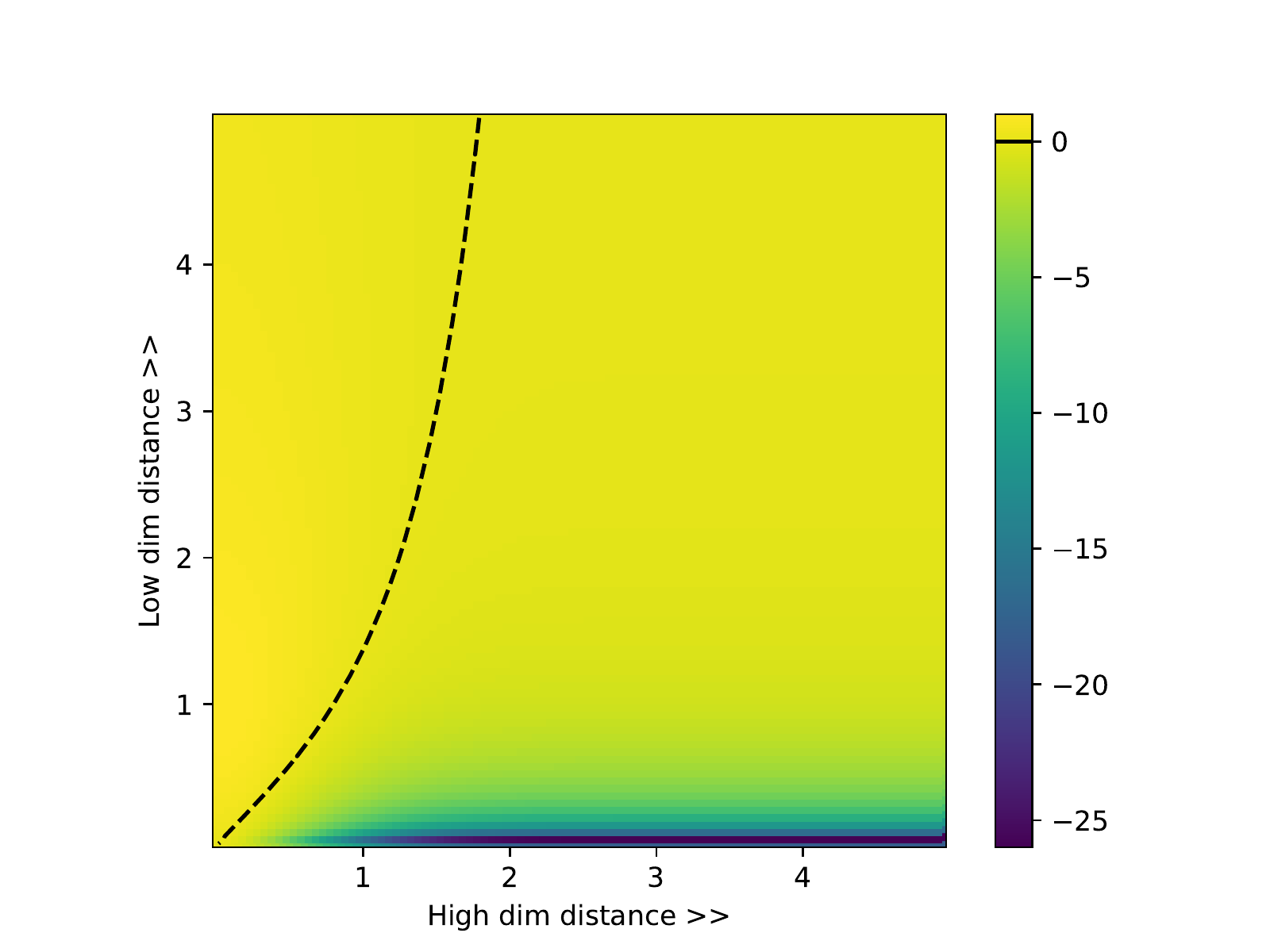} & 
        \includegraphics[width=\linewidth]{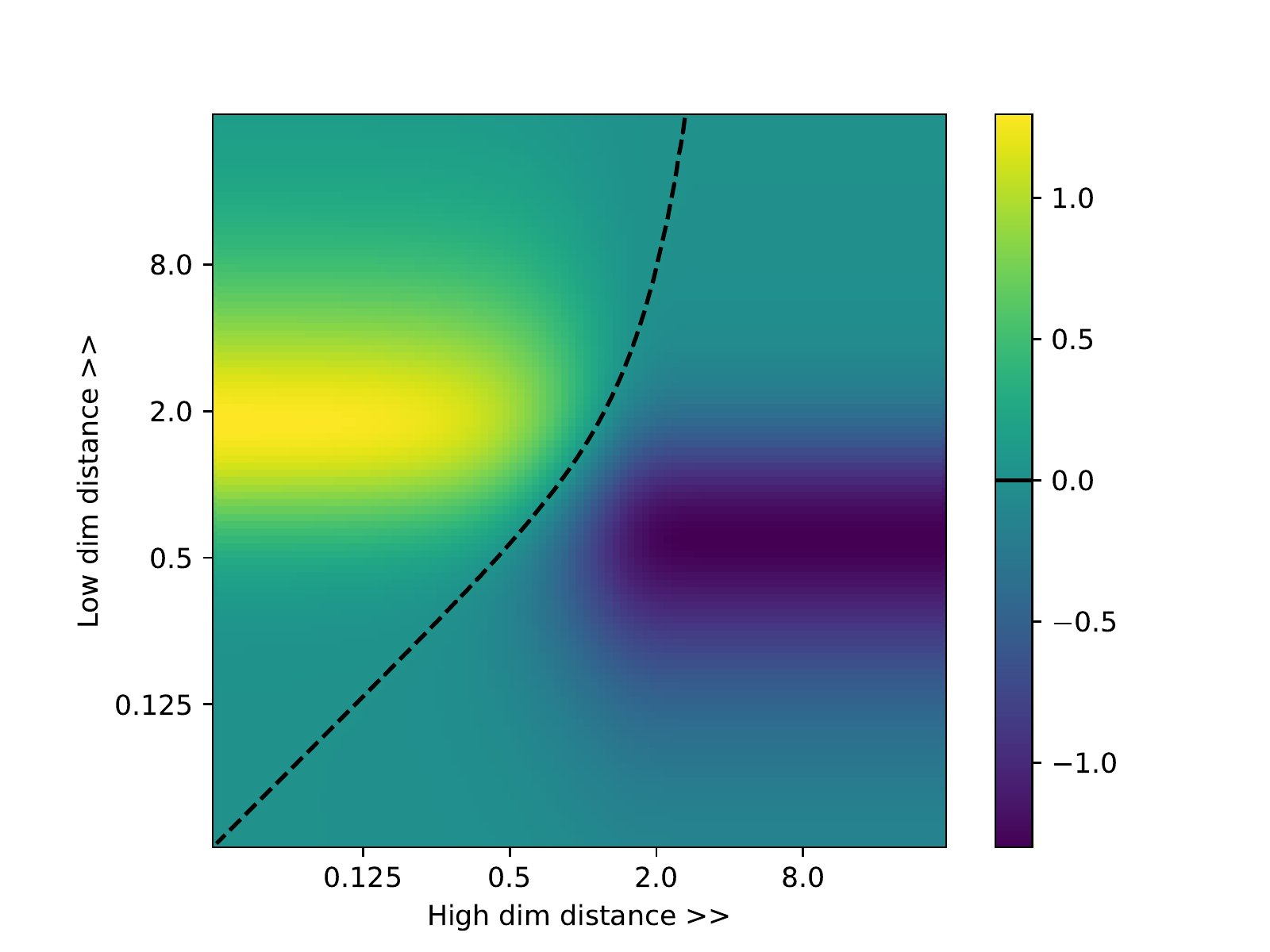} & 
        \includegraphics[width=\linewidth]{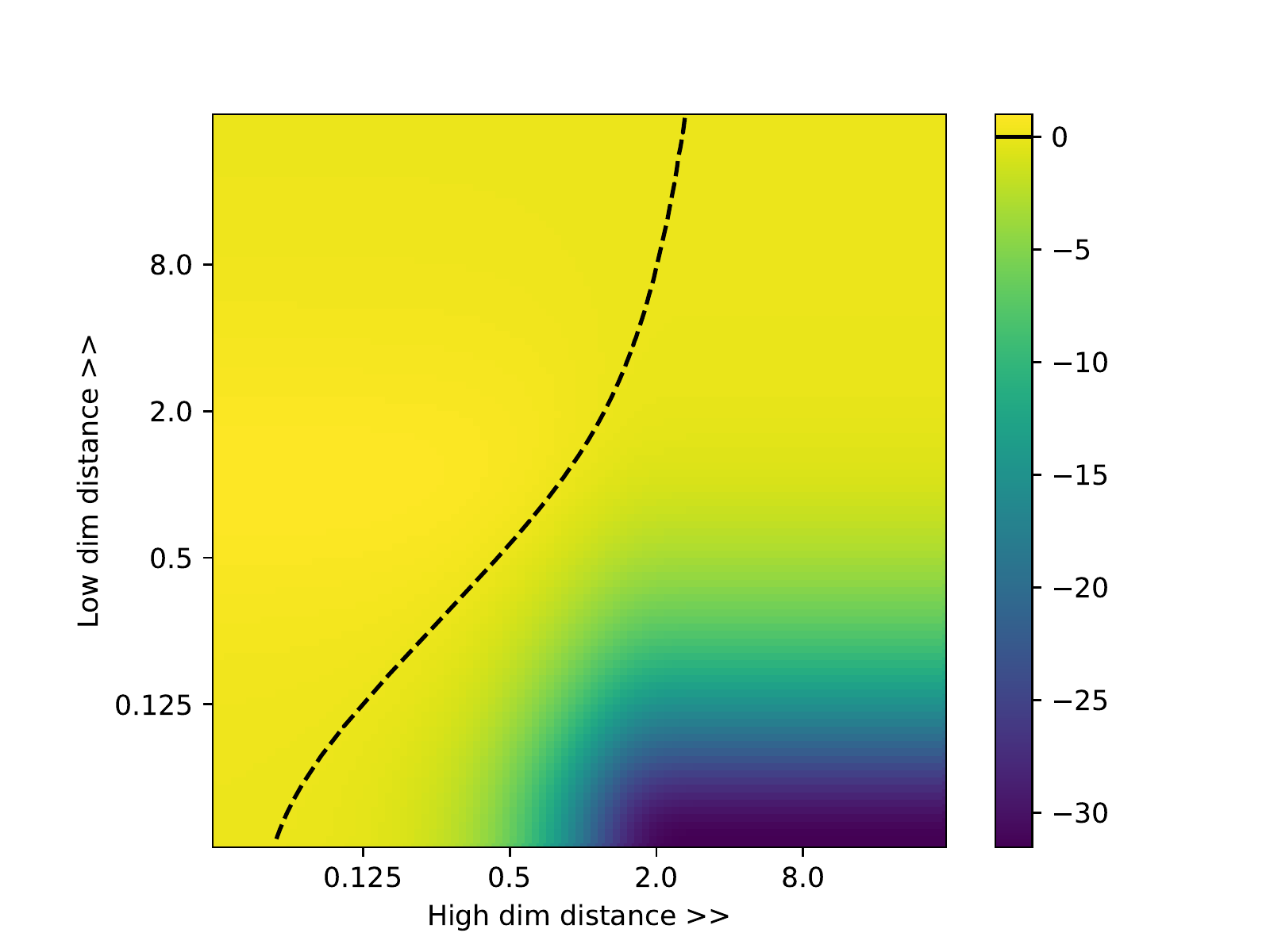}\\
        Frobenius Norm &
        \includegraphics[width=\linewidth]{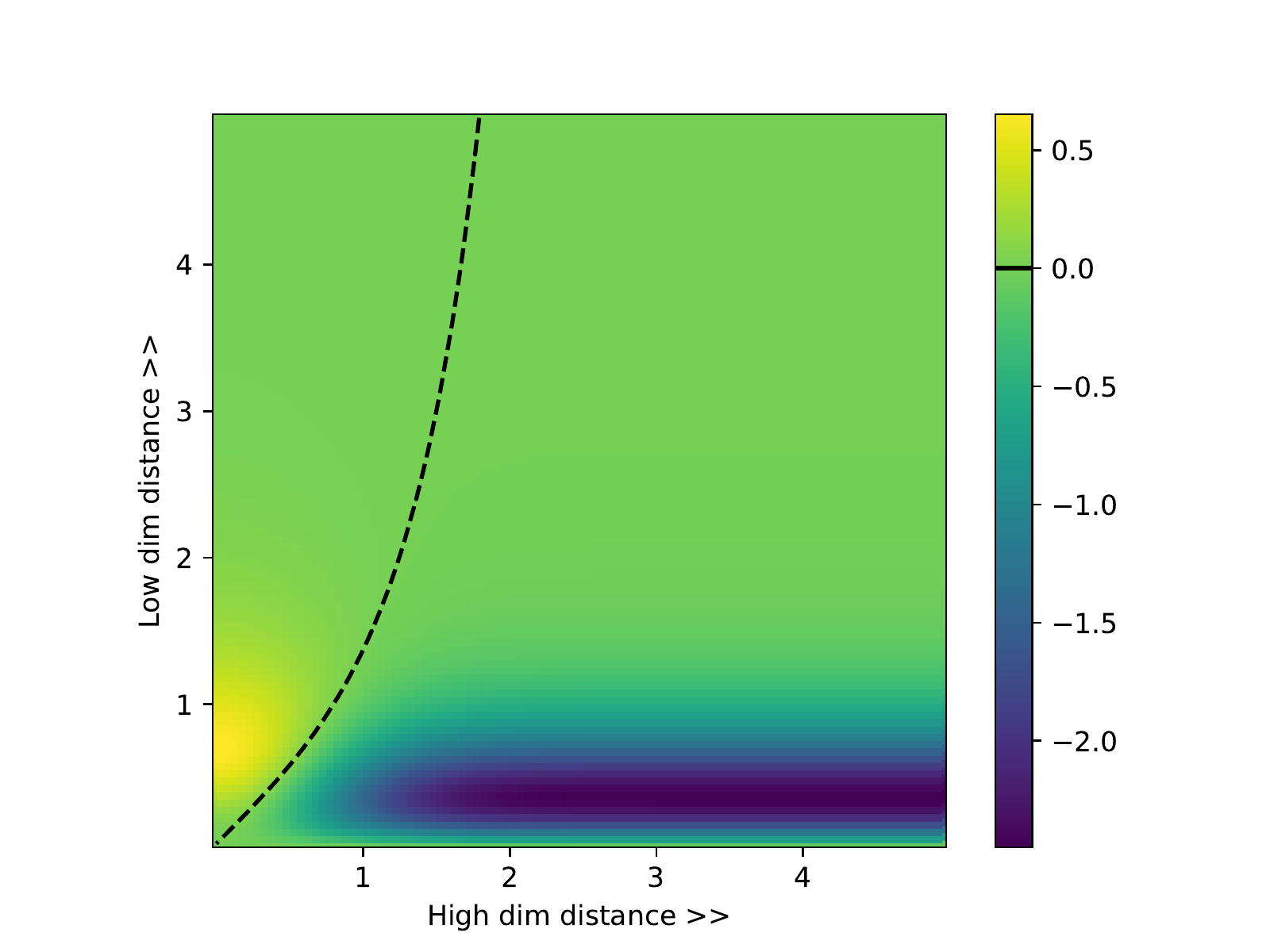} & 
        \includegraphics[width=\linewidth]{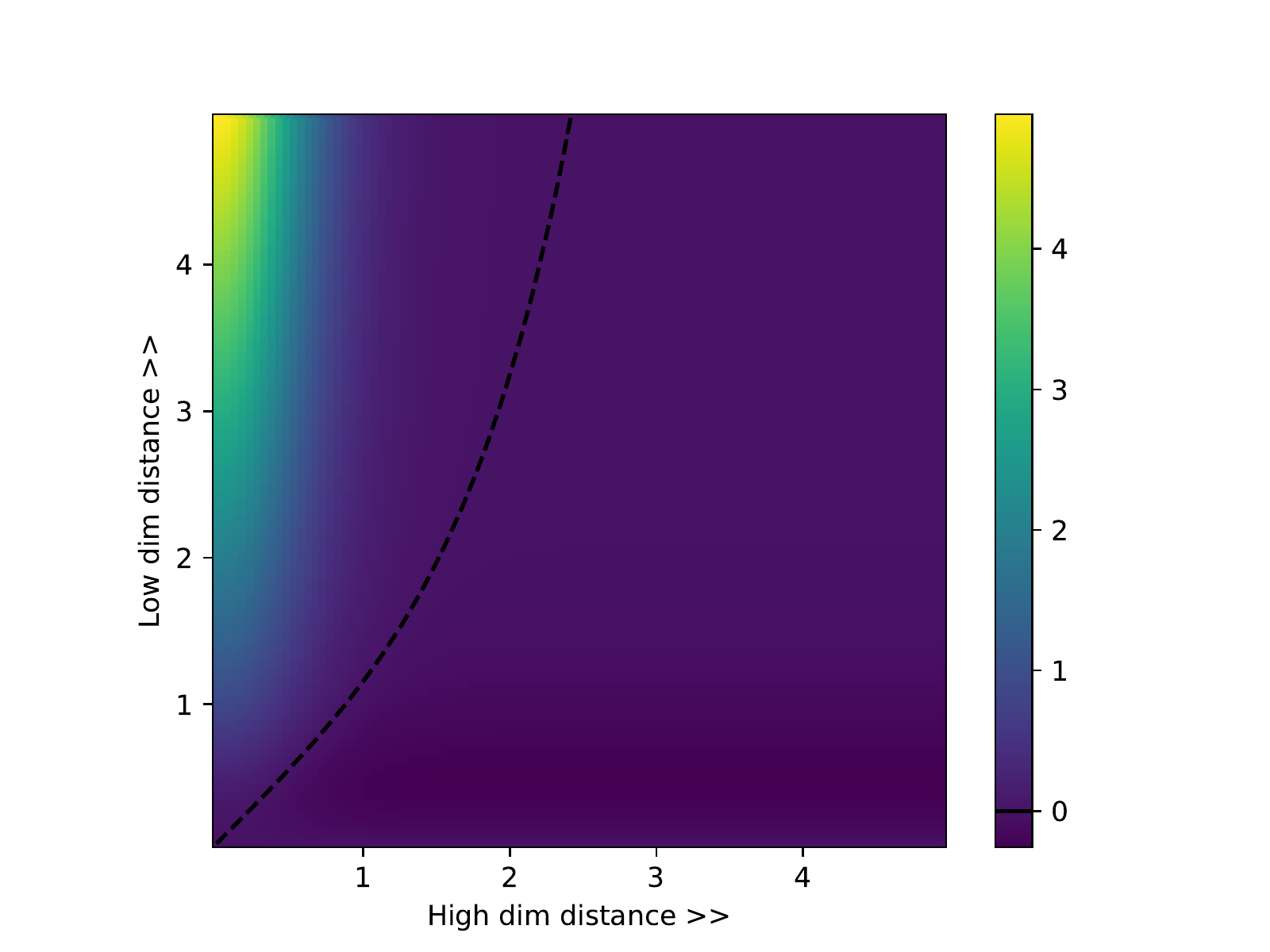} & 
        \includegraphics[width=\linewidth]{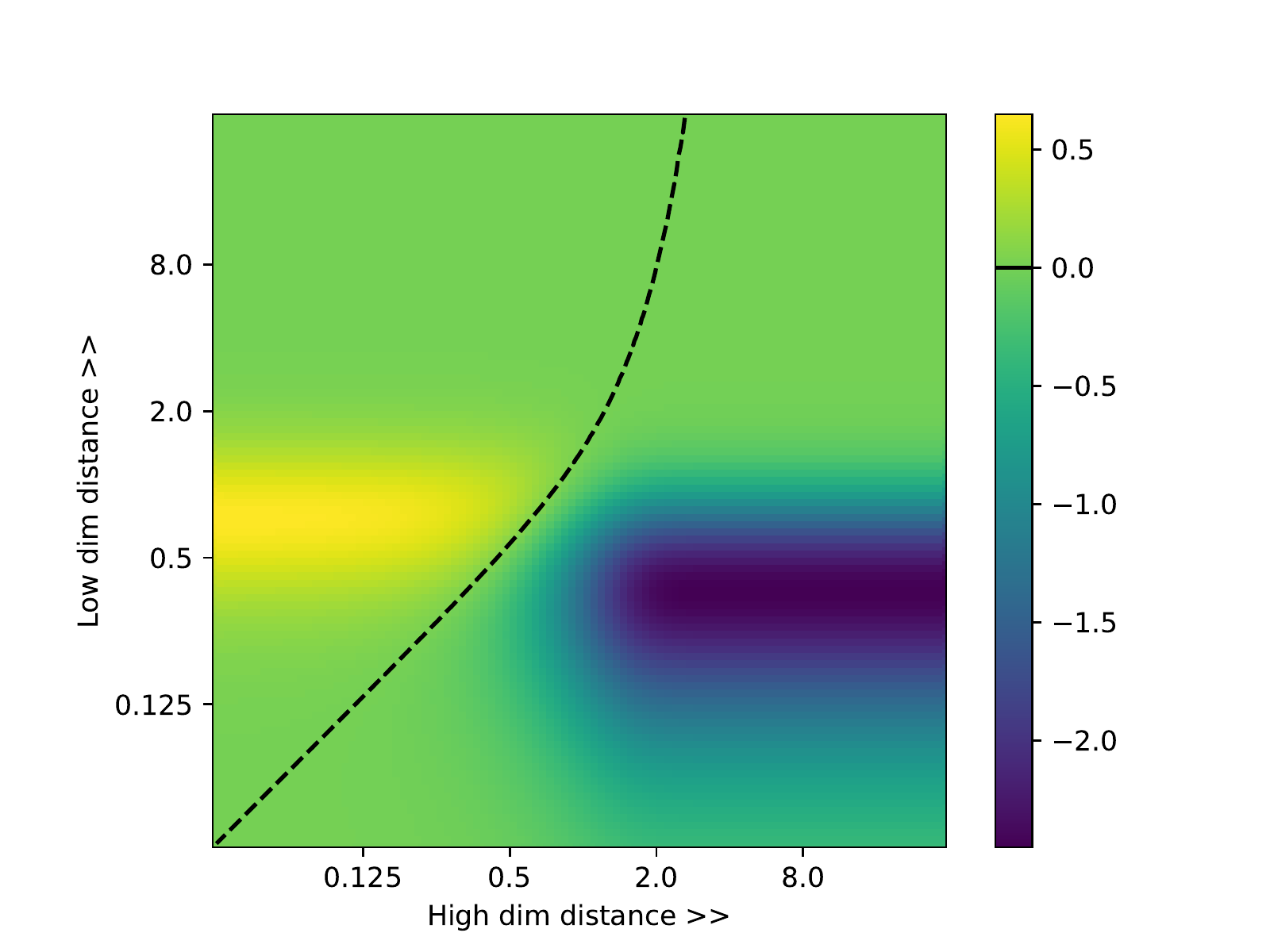} & 
        \includegraphics[width=\linewidth]{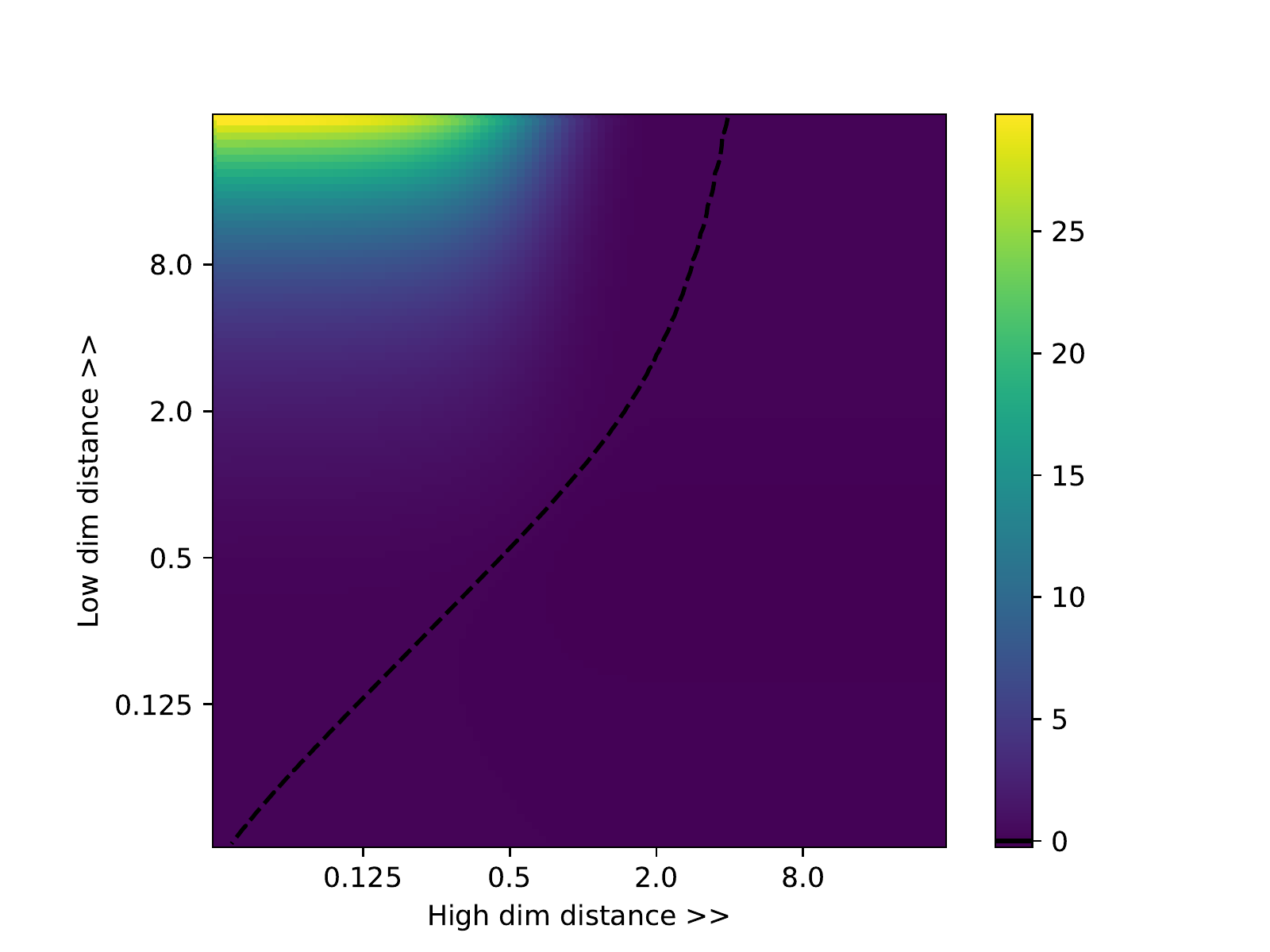}\\
    \end{tabular}
\caption{Gradient relationships between high-dim and low-dim distances for TSNE and UMAP. The dotted line represents the locations of magnitude-$0$ gradients. Since the X and Y axes are the same in every image, if two images have $0$-magnitude gradients then this solution satisfies any of the cost functions. Thus, TSNE and UMAP under both the KL-divergence and Frobenius norm share many of their minima. The top-left image is a recreation of the original gradient plot in \cite{van2008visualizing}.}
\label{grad_plots}
\vspace{-3mm}
\end{figure*}
\begin{figure*}[h]
    \begin{tabular}{cccc}
        \includegraphics[width=.22\linewidth]{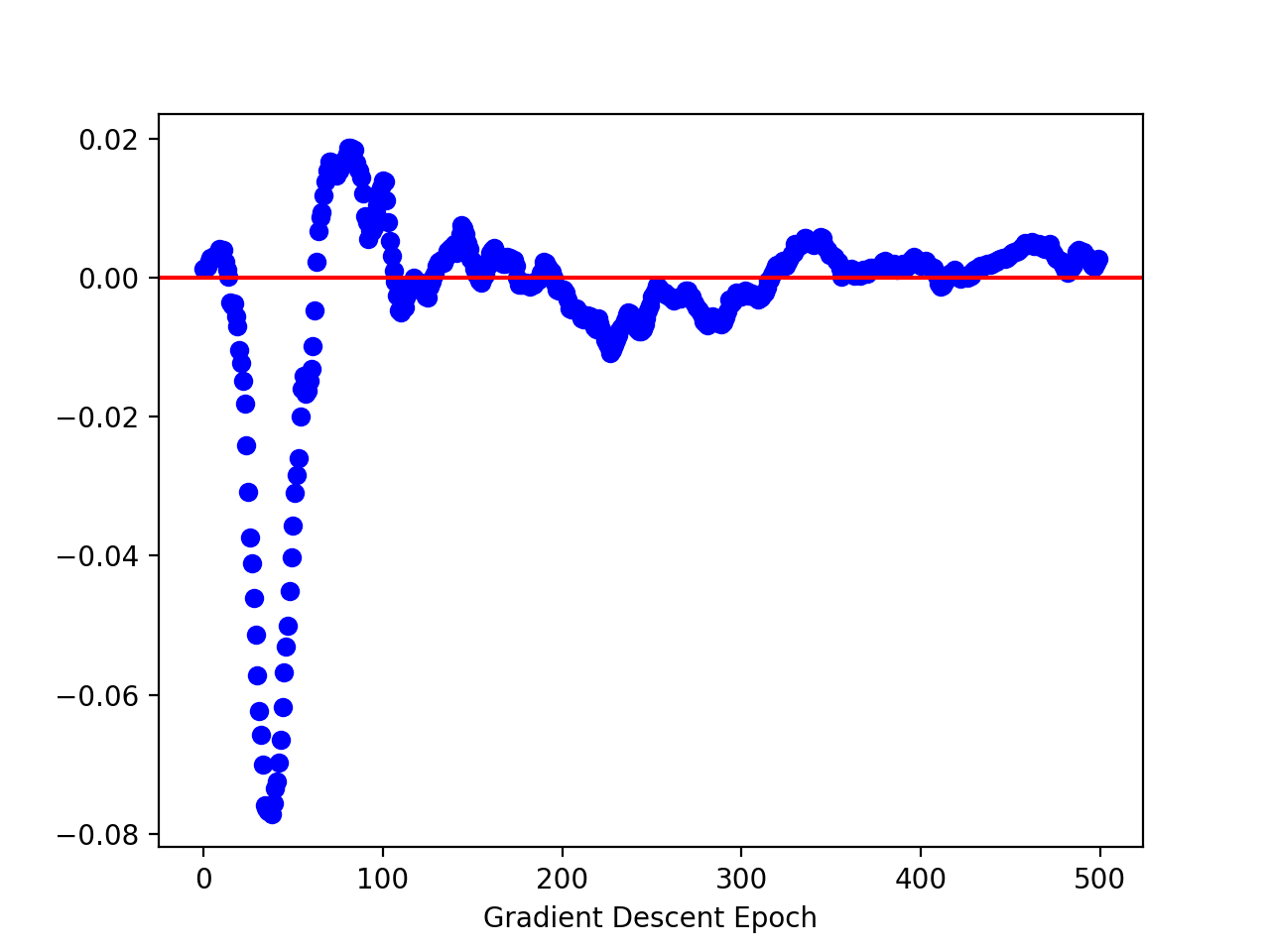} & 
        \includegraphics[width=.22\linewidth]{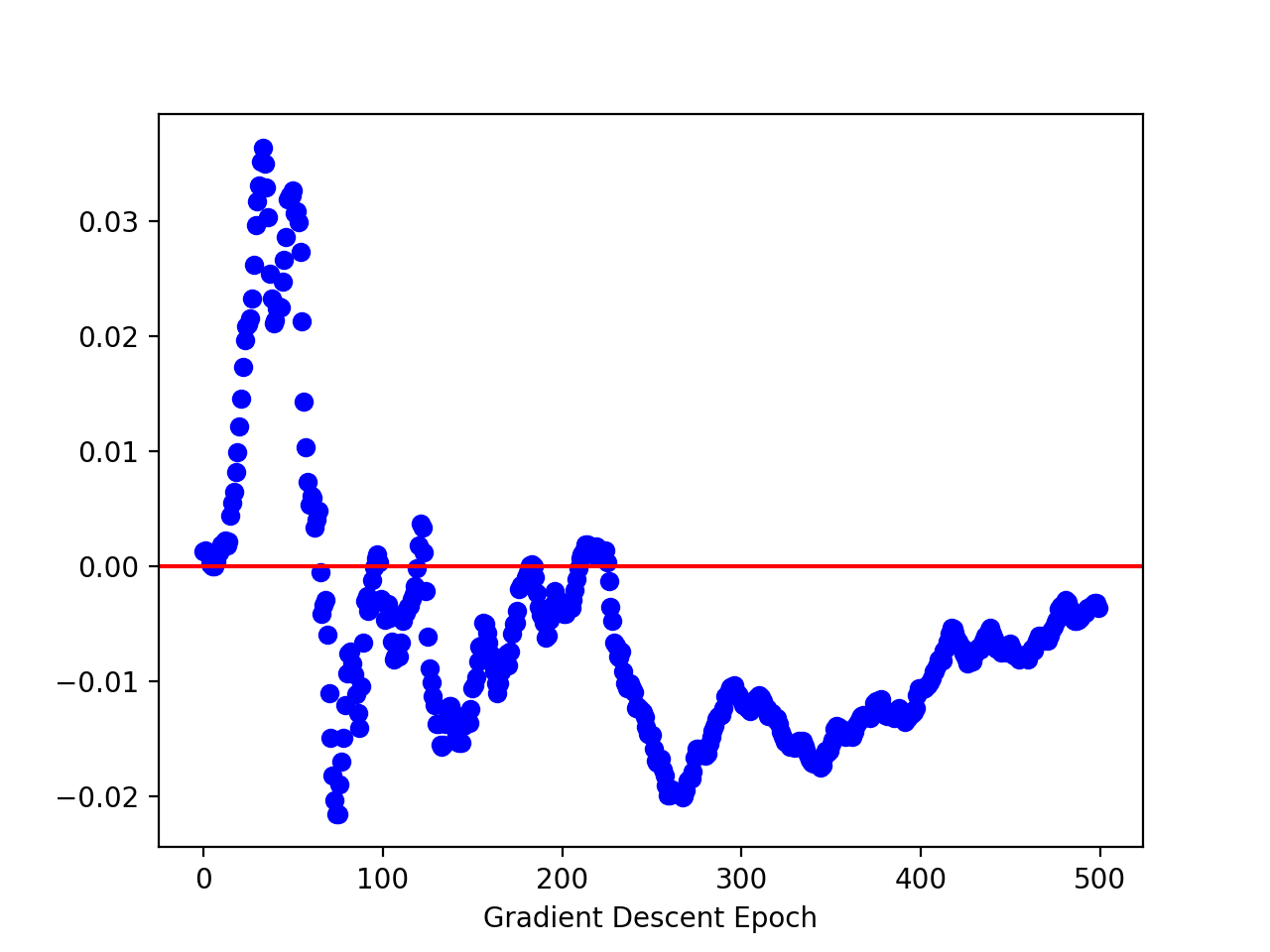} & 
        \includegraphics[width=.22\linewidth]{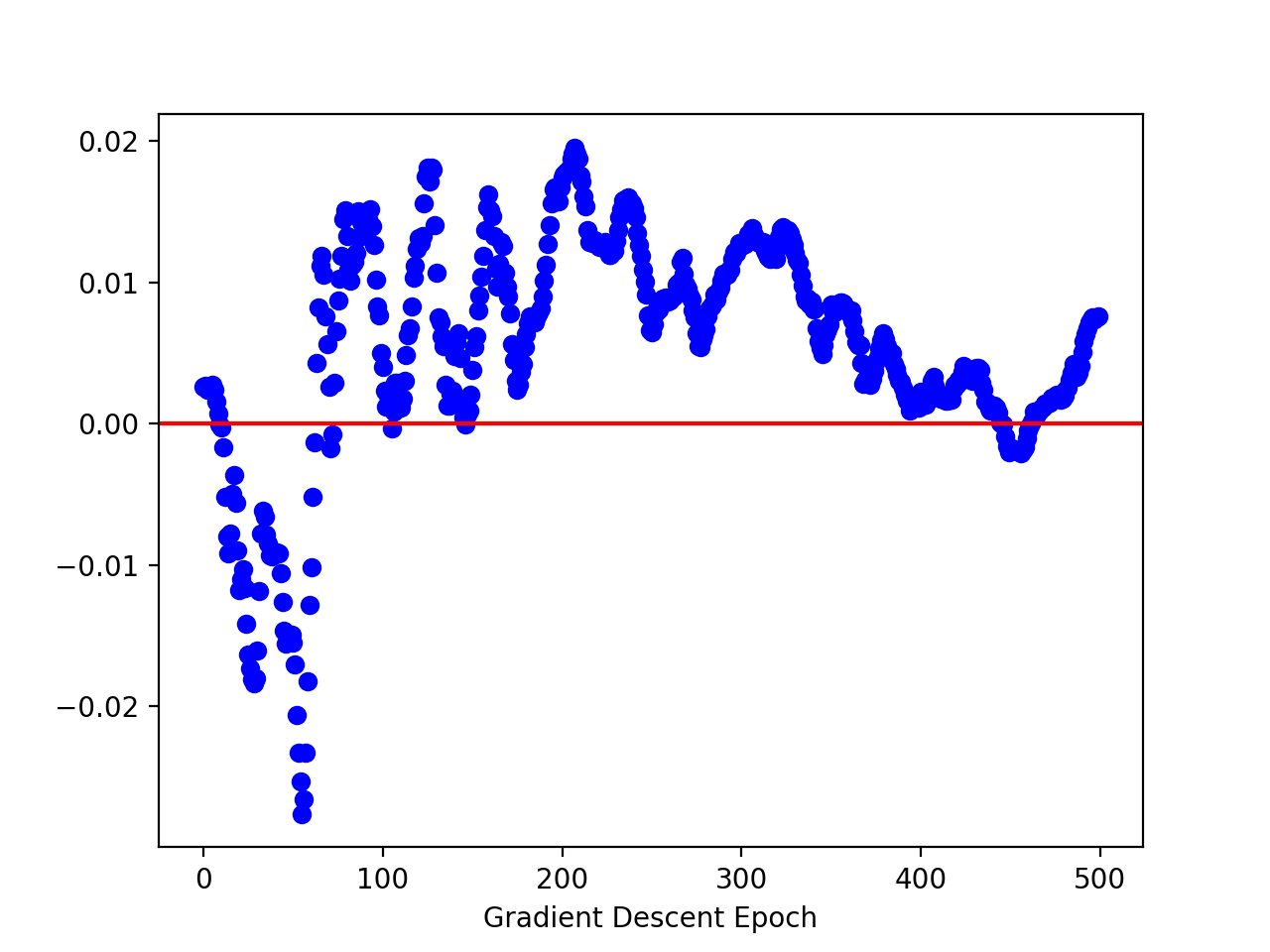} & 
        \includegraphics[width=.22\linewidth]{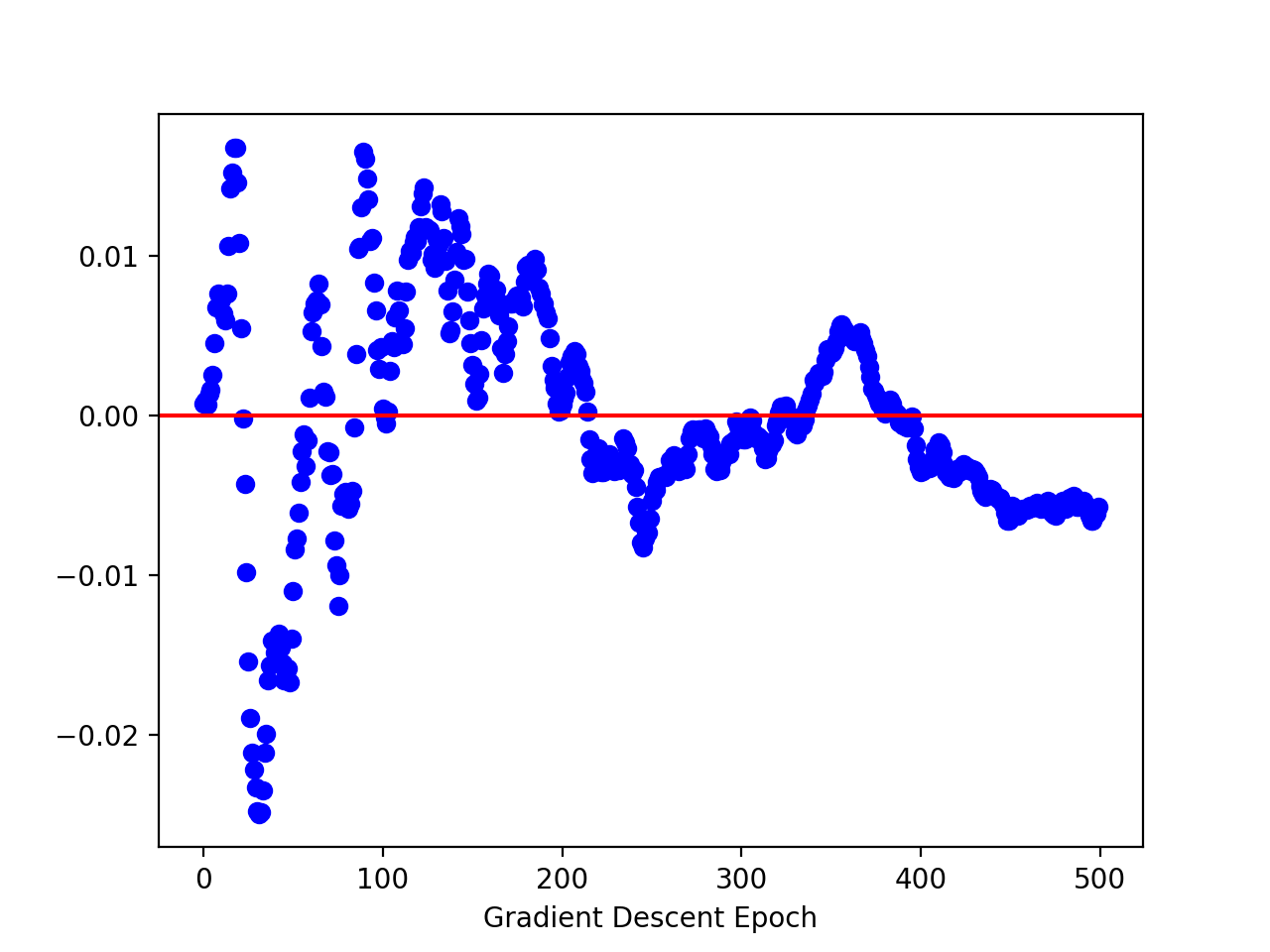}\\
        MNIST & Fashion MNIST & COIL100 & Swiss Roll \\
    \end{tabular}
\caption{Average angle in radians between repulsive forces calculated by TSNE Barnes-Hut trees and UMAP's point sampling over the course of training. This verifies that the gradients between the normalized and unnormalized optimization criteria consistently point in the same direction.}
\label{grad_agreement}
\vspace{-3mm}
\end{figure*}

\subsection{The Principal Difference}
Given the above gradient derivations, it may be surprising to know that TSNE evaluates $\bigO(n^2)$ repulsive forces during an epoch while UMAP only calculates $\bigO(n)$ of them. Both of their gradients are defined as the sum over all $i \neq j$ so one should expect that they perform similar numbers of repulsions during optimization. As can be seen in Figure~\ref{repulsion_vis}, however, TSNE estimates all $n-1$ repulsive forces acting on each point while UMAP simply sums the repulsions from a constant number of points. TSNE estimates the $n-1$ repulsions by fitting a Barnes-Hut tree \cite{barnes1986hierarchical} to $Y$ at every gradient descent step; this allows groups of distant points to be treated as a single point with larger weight. Importantly, this additional step induces the principal speed discrepancy between the two approaches, with TSNE spending the majority of the time fitting the Barnes-Hut trees in low-dimensional space. Does this mean that TSNE could employ only a constant number of repulsions as well? If not, what changes should be made such that TSNE can optimize at UMAP's speeds?

We make the argument that the change in normalization is the fundamental difference between the two algorithms in terms of both their speeds and their resulting embeddings. First, note that the UMAP repulsive force in Equation~\ref{umap_rep} is
inversely quadratic with respect to the low-dimensional distance, leading to extreme repulsions between points that are too close in the low-dimensional space. This comes directly as a consequence of the $\log(1 - q_{ij})$ term in the unnormalized KL divergence, since

\begin{align*}
\dfrac{\partial \log(1 - q_{ij})  }{\partial y_i} &= \dfrac{1}{1 - q_{ij}} \cdot \dfrac{\partial q_{ij}}{\partial y_i} \\
&= \dfrac{1}{1 + \dfrac{1}{1 - a \|y_i - y_j\|^{2b}}} \cdot \dfrac{\partial q_{ij}}{\partial y_i} \\
&= \dfrac{1 + a \|y_i - y_j\|^{2b}}{a \|y_i - y_j\|^{2b}} \cdot \dfrac{\partial q_{ij}}{\partial y_i},
\end{align*}
where the term in the numerator will cancel after expanding $\partial q_{ij} / \partial y_i$.

This inverse relationship to the distance is inherently unstable and is handled computationally by adding an $\epsilon$ additive term. Nonetheless, UMAP still receives unwieldy repulsion values and has to additionally manage this by clipping the gradients and using non momentum-based gradient methods.
This difference in magnitudes can also be seen in Figure~\ref{grad_plots}, where we plot the gradient magnitude as a function of high- and low-dimensional distances (for a constant number of points). Notice that the repulsive values of UMAP under the KL-divergence far exceed those in TSNE. This means that applying gradient amplification does not interact well with unnormalized gradients, as the momentum term will latch on to these extreme repulsions and destabilize the optimization process. 

Beyond this edge case, the normalization's more important effect is on the magnitude of the average gradient.
We see that TSNE's attractive and repulsive forces only cancel \textit{one} of the two normalization factors that are present. Namely, the $Z = \sum_{k \neq l} (1 + \|y_k - y_l\|^2)^{-1}$ cancels the normalization in $q_{ij}$, but the normalization in the remaining $p_{ij}$ and $q_{ij}$ term still remains. In essence, this means that each TSNE attractive and repulsive force is inversely proportional to the number of points. UMAP, on the other hand, has no such normalization factor. In practice, this creates the effect that UMAP's gradients are at least a factor of $n$ stronger than the corresponding TSNE ones and explains the seemingly contradictory repulsion methodologies where UMAP calculates a constant number of repulsions per point while TSNE estimates all $n-1$.

As discussed in the above paragraphs, the normalization goes hand-in-hand with the gradient-descent strategy. Intuitively, minimizing TSNE's loss without gradient amplification does not move the points at all while minimizing UMAP's loss with gradient amplification is highly unstable. For this reason, normalizing the pairwise similarity matrices will imply performing gradient amplification going forward (and vice versa).


    
    
    
    
    
    
    


\newcommand{\dcell}[1]{\small{\it\color{gray}\makecell{#1}}}

\begin{table*}[thb]
    \newcolumntype{C}{>{\centering\arraybackslash}X}
    \newcolumntype{H}{>{\setbox0=\hbox\bgroup}c<{\egroup}@{}}
    \begin{tabularx}{\textwidth}{Ccccccc}
    & \textbf{Normalization} & \textbf{Initialization} & \textbf{Distance function} & \textbf{Symmetrization} & \textbf{Sym Attraction} & \textbf{Scalars} \\
     & \dcell{$P, Q$  normalized} & \dcell{$Y$ initialization} & \dcell{High-dim \\ distances calculation} & \dcell{Setting {$p_{ij} = p_{ji}$}} & \dcell{Attraction$(y_i, y_j($\\ applied to both} & \dcell{Values for \\ $a$ and $b$} \\
    \midrule
    
    TSNE & Yes & Random & $d(x_i, x_j)$ & $(p_{i|j} + p_{j|i})/2$ & No & $a = 1$, $b = 1$\\
    
    UMAP & No & Lapl. Eigenmap & \makecell{$d(x_i, x_j) - \min_k d(x_i, x_k)$} & $p_{i|j}{+}p_{j|i}{-}p_{i|j}p_{j|i}$ & Yes & Grid search \\
    \bottomrule
    \end{tabularx}
    \caption{A full list of the differences between TSNE and UMAP that we analyze in Tables~\ref{irrelevant-metrics}, \ref{irrelevant-mnist} and ~\ref{irrelevant-fashion-mnist}. }
    \label{differences_table}
\end{table*}
\section{\ourmethod} 
\label{uniform}

We now present \ourmethod  --- an algorithm that can recreate both TSNE and UMAP embeddings at speeds faster than available UMAP methods. The key observation is noticing that the only necessary change between TSNE and UMAP is the normalization. Given this context, our algorithm follows the UMAP optimization procedure except that we (1) replace the \textit{scalar sampling} by iteratively processing attractions/repulsions and (2) perform gradient descent outside of the gradient-collection for-loop. The first change accommodates the gradients under normalization since the normalized repulsive forces do not have the $1 - p_{ik}$ term that UMAP samples proportionally to. The second change allows for performing momentum gradient descent for faster convergence in the normalized setting. Due to these algorithmic changes, \ourmethod is particularly amenable to parallelization -- allowing us to calculate the objective functions with fewer approximations.

Given that we follow the UMAP optimization procedure, \ourmethod defaults to recreating the UMAP embeddings. In the case of replicating TSNE, we can simply normalize the $P$ and $Q$ matrices and scale the learning rate by $n/k$. This scaling allows us to simulate the gradient magnitudes from all $n-1$ repulsions while only calculating a constant number of them, essentially circumventing the need to fit a Barnes-Hut tree during each epoch. Intuitively, this only works if the angle between the TSNE and UMAP repulsions is consistently small. We experimentally validate this to be the case by plotting the average angle between the TSNE Barnes-Hut repulsions and the UMAP sampled repulsions in Figure~\ref{grad_agreement}. Across datasets, we see that the direction of the repulsion remains consistent throughout the optimization process, implying that we can safely calculate the repulsion with respect to a constant number of points and then scale its magnitude accordingly. Going forward, we refer to \ourmethod as \ourmethodU if it is in the unnormalized setting and as \ourmethodN if it is in the normalized setting. Thus, \ourmethodU has similar gradients to UMAP and \ourmethodN has similar gradients to TSNE; we refer the reader to Figure~\ref{grad_plots} for a depiction of these gradient magnitudes across distances and loss functions. For smoother convergence, we toggle the gradient amplification with the normalization as discussed in Section~\ref{grad_calc_sec}.

By implementing the UMAP optimization protocol such that the normalization can be toggled, we are free to choose whichever options we like across the other settings. As will be shown in Section \ref{results}, the remaining algorithmic choices in Table \ref{differences_table} are negligible across datasets and metrics. Thus, we can choose whichever are most efficient. \ourmethod therefore defaults to TSNE's asymmetric attraction and $a, \; b$  scalars along with UMAP's distance-metric, initialization, nearest neighbors, and $p_{ij}$ symmetrization. 

As previously mentioned, most values of $p_{ik}$ are unavailable to us during optimization since we only calculated the $P$ matrix for nearest neighbors in the high-dimensional space. We standardize the UMAP approximation of $p_{ik} \approx p_{ij}$ by setting $p_{ik} = \bar{p}_{ij} \; \forall \; p_{ik}$, where $\bar{p}_{ij}$ is the mean value of $P$ (for known values in $P$).

\subsection{\ourmethod with the Frobenius Norm}
Recall that the $1 - p_{ik}$ scalar is a direct consequence of the KL-divergence in the unnormalized setting. We must then replace the KL divergence with another loss function if we want to avoid unnecessary estimations. Although the KL divergence has been used across gradient-based DR methods \cite{mcinnes2018umap, van2008visualizing, van2014accelerating, linderman2019fast, tang2016visualizing}, we find that one can also optimize the squared Frobenius norm without sacrificing quality:
\[ \mathcal{L}(X, Y) = \sum_{i, j} (p(x_i, x_j) - q(y_i, y_j))^2 \]

This presents us with the following attractive and repulsive forces acting on point $y_i$
\begin{align*}
    \mathcal{A}_i^{frob-tsne} &= -4 \sum_{j} p_{ij} Z (q_{ij}^2 + 2q_{ij}^3) (y_i - y_j) \\
    \mathcal{R}_i^{frob-tsne} &= 4 \sum_{j} Z( q_{ij}^3 + 2q_{ij}^4) (y_i - y_j) \\
    &\\
    \mathcal{A}_i^{frob-umap} &= -4 \sum_{j} p_{ij} q_{ij}^2 (y_i - y_j) \\
    \mathcal{R}_i^{frob-umap} &= 4 \sum_{j} q_{ij}^3 (y_i - y_j)  
\end{align*}

Although the squared Frobenius norm in Tables~\ref{irrelevant-mnist},~\ref{irrelevant-metrics}~and~\ref{irrelevant-metrics-other-data} loses the KL-divergence's probabilistic interpretation and has a significantly different gradient formula, it yields similar embeddings. Nevertheless, the Frobenius norm shares the same KL-divergence's minimum (namely $p_{ij} = q_{ij}$) and preserves the property that a majority of the gradient space has magnitude zero (deep blue area in Table~\ref{grad_plots}). On the other hand, the Frobenius norm is easier to optimize as it avoids the $1 - p_{ik}$ scaling factor. Lastly, the Frobenius norm gradients\footnote{We wrote the gradients under the assumption that $a = b = 1$, although they are simple to calculate in the general setting as well.} provide a convenient function for the attractions and repulsions in the unnormalized setting. However, we do not default to minimizing the Frobenius norm in \ourmethod as we aim to produce TSNE and UMAP embeddings with minimal changes.


\section{Results} \label{results}

In our experiments we validate a number of claims and show that our \ourmethod not only accurately reproduces the results of both TSNE and UMAP, but also consistently outperforms the best versions of each algorithm in terms of speed. The experiments proceed as follows. First, Section~\ref{ssec:irrelevant_alg_params} confirms that most of the hyperparameters identified in Table~\ref{differences_table} have a negligible effect on both TSNE and UMAP. Second, Section~\ref{ssec:norm_results} confirms that the normalization is the primary difference between TSNE and UMAP. This leads to the conclusion that normalization allows \ourmethod to toggle between the outputs of the two algorithms. Finally, Section~\ref{ssec:alg_comparison_results} presents an efficiency comparison and shows the advantages of \ourmethod. Specifically, we highlight that our method runs at least as fast as our optimized version of UMAP, implying that it significantly outperforms the traditional UMAP library as well as all available TSNE methods. We also show, quantitatively and qualitatively, that the embeddings obtained by \ourmethod are indistinguishable from those that TSNE and UMAP would produce. We run the experiments on an Intel Core i9 10940X 3.3GHz 14-Core processor and 256 GB of RAM.

\spara{Metrics.} We employ standard measures for clustering and dimensionality reduction to quantitatively evaluate the embeddings. 

We report the \emph{kNN-accuracy}, i.e., the accuracy of a k-NN classifier, to assert that objects
of a similar class remain close in the embedding. Assuming that intra-class distances are smaller than inter-class distances in the high-dimensional
space, a high kNN-accuracy implies that the method effectively preserves similarity during dimensionality reduction. Unless stated otherwise, we
choose $k = 100$ in-line with prior work~\cite{mcinnes2018umap}.

We study embedding consistency by evaluating KMeans clustering on datasets that have class labels. We report cluster quality in terms of \emph{homogeneity} and \emph{completeness}~\cite{rosenberg2007v}.
Homogeneity is maximized by assigning \textit{only} data points of a single class to a single cluster.
Completeness serves as homogeneity's symmetric counterpart and is maximized by assigning \textit{all} data points of a class to a cluster.
Both metrics lie within the range $[0, 1]$.
A good clustering solution must then balance achieving high homogeneity with high completeness, as either metric can be trivially set to 1 at the expense of the
other being 0. For brevity, we report the \emph{V-measure}, the average between the homogeneity and completeness. This metric simply relies on the labeling and does not take the point locations into account. As such, it is invariant to the biases inherent in KMeans and serves as a more objective measure than KMeans loss.

Additionally, we note that dimensionality reduction methods are often judged by their qualitative ability to make good dataset visualizations. This is difficult to
quantify, so we use our best judgement when saying that two embeddings with comparable metrics ''look`` similar.

\spara{Datasets.} In our evaluation, we use six standard datasets used in dimensionality reduction. In particular, we employ the popular MNIST~\cite{lecun-mnisthandwrittendigit-2010}, Fashion-MNIST~\cite{xiao2017fashion}, CIFAR~\cite{krizhevsky2009learning}, Coil~\cite{nene1996columbia} image datasets, the Google news~\cite{mikolov2013efficient}\textbf{} text embedding dataset and the Swiss Roll synthetic dataset. Table~\ref{tbl:datasets} reports the main data characteristics. Most of the datasets assume the Euclidean distance metric with the sole exception being the Google News dataset using the cosine distance.

\begin{table}[H]
\centering
\setlength{\tabcolsep}{2pt}
\newcolumntype{C}{>{\centering\arraybackslash}X}
\begin{tabularx}{\linewidth}{XrrrX}
\toprule
\textbf{Dataset} & $n$ & $c$ & $D$ & Type\\ 
\midrule
MNIST \cite{lecun-mnisthandwrittendigit-2010} &  60\,000  & 10 & 784 & Images \\
Fashion-MNIST \cite{xiao2017fashion} &  60\,000  & 10 & 784 & Images\\
CIFAR-10 \cite{krizhevsky2009learning} &  60\,000  & 10 & 3\,072 & Images\\
Coil-100 \cite{nene1996columbia} &  7\,200  & 100 & 49\,152 & Images \\
\midrule
Swiss Roll &  5\,000  & -- & 3 & Synthetic\\
Google News \cite{mikolov2013efficient} &  350\,000  & -- & 300 & Text embeddings\\
\bottomrule
\end{tabularx}
\caption{Datasets used in our evaluation, $n$ samples, $c$ classes, $D$ dimensions. Swiss Roll and Google News are unlabeled.}
\label{tbl:datasets}
\vspace{-4mm}
\end{table}


\begin{table*}[!htb]

    \centering
    \begin{tabularx}{\textwidth}{c*{6}{c}}
    & \multicolumn{6}{c}{\textit{Swapped setting}}\\
    \cline{2-7}
    \textit{Original setting} & Frobenius & Initialization & Pseudo-distance & Symmetrization & Sym attraction & Scalars\\

    \midrule
    \includegraphics[width=.123\linewidth]{outputs/mnist/tsne/default_embedding.png}&
    \includegraphics[width=.123\linewidth]{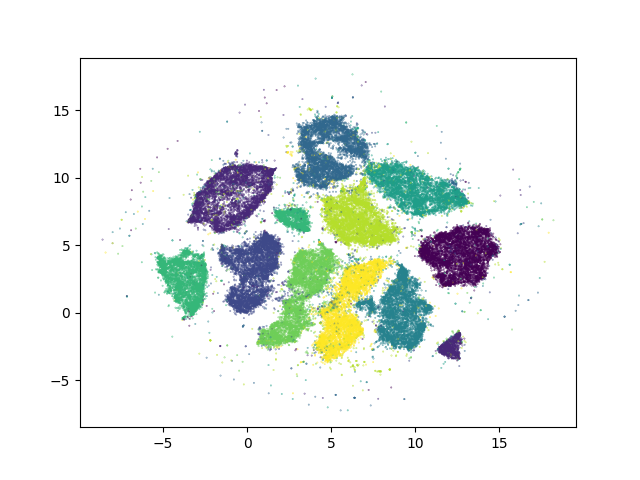}&
    \includegraphics[width=.123\linewidth]{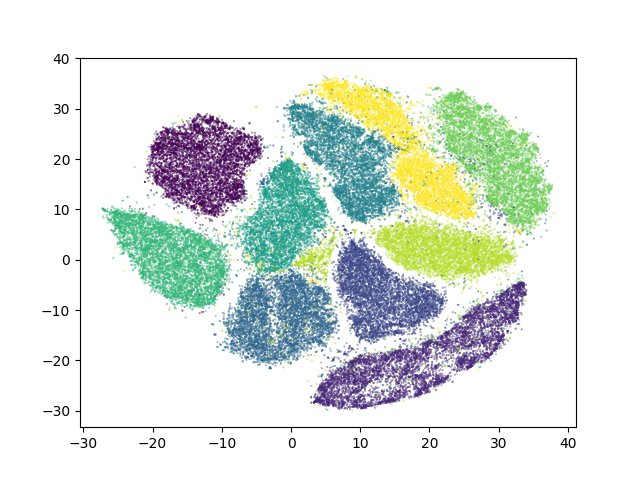}&
    \includegraphics[width=.123\linewidth]{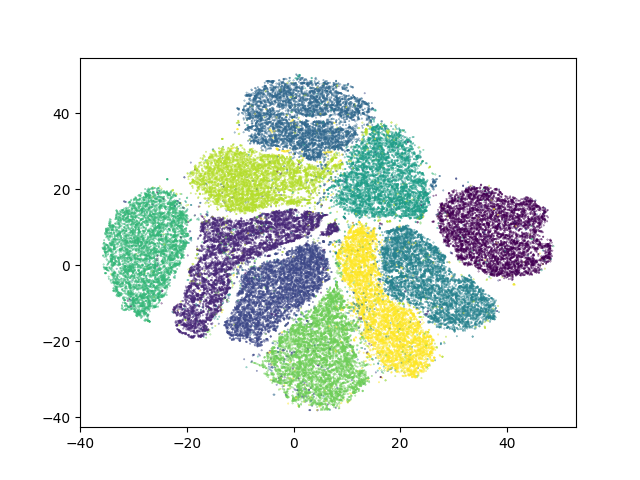} & 
    \includegraphics[width=.123\linewidth]{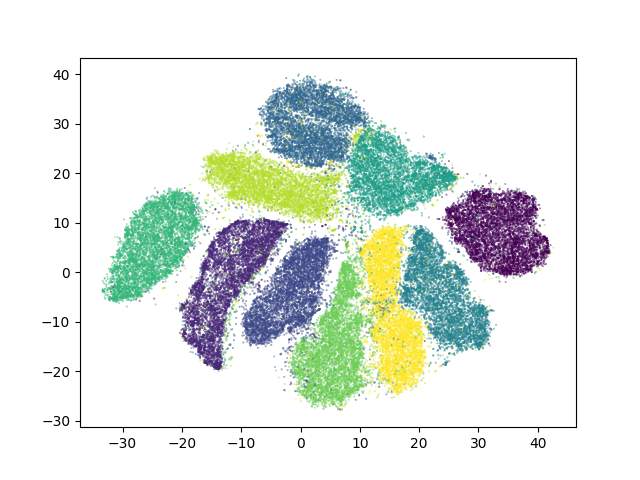}&
    \includegraphics[width=.123\linewidth]{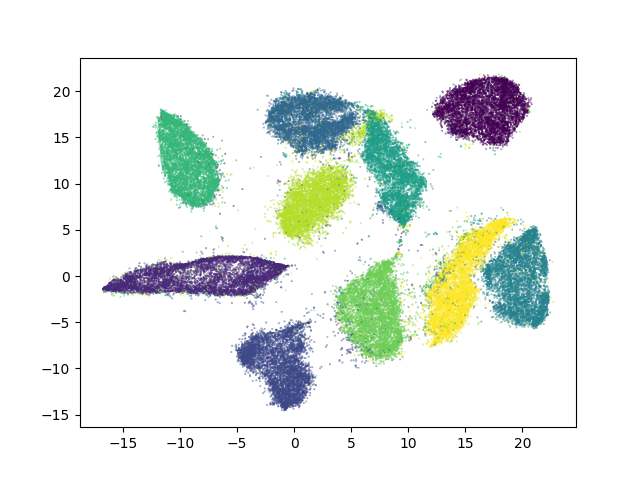}&
    \includegraphics[width=.123\linewidth]{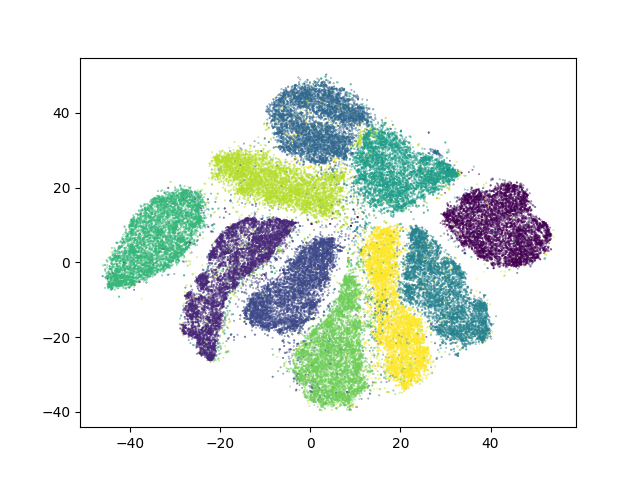}\\

    \includegraphics[width=.123\linewidth]{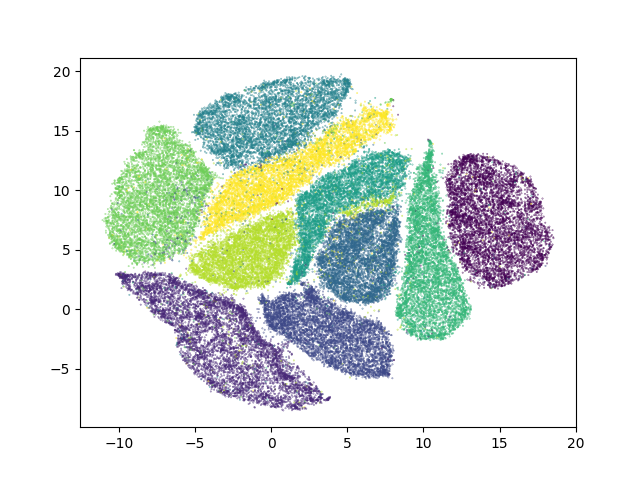}&
    \includegraphics[width=.123\linewidth]{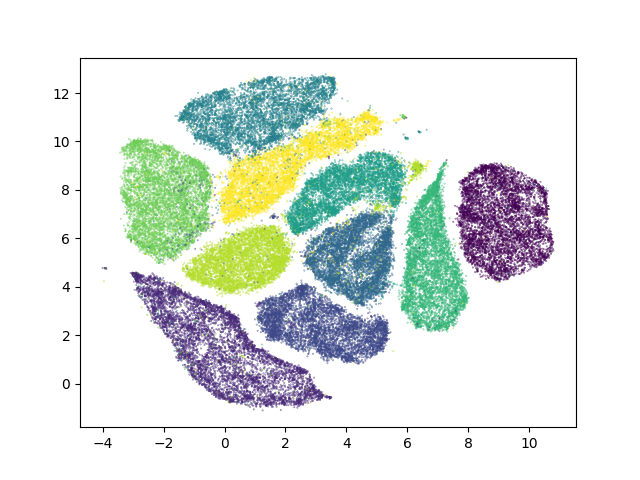}&
    \includegraphics[width=.123\linewidth]{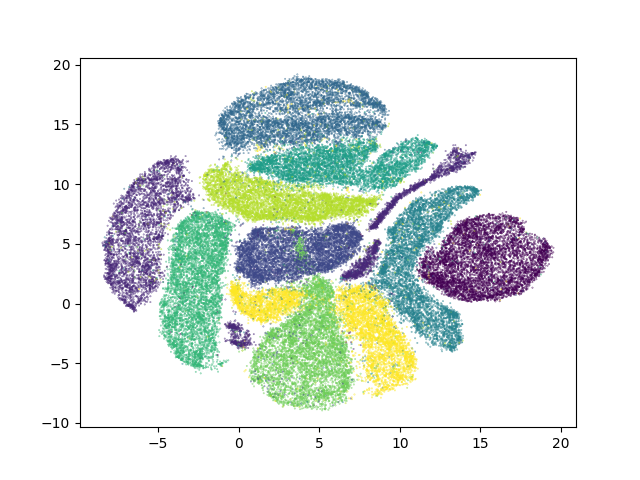}&
    \includegraphics[width=.123\linewidth]{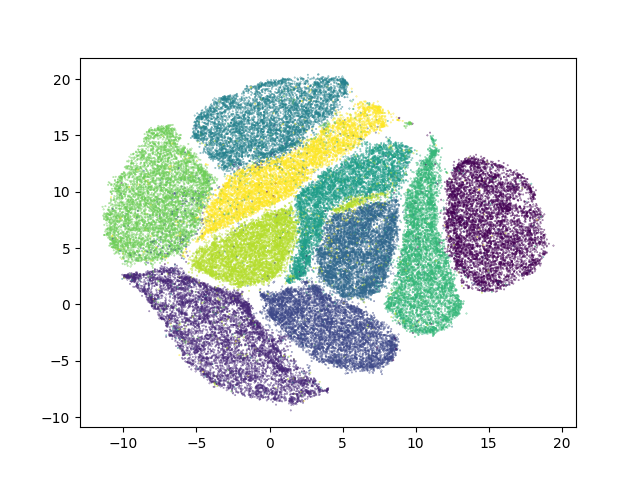}& 
    \includegraphics[width=.123\linewidth]{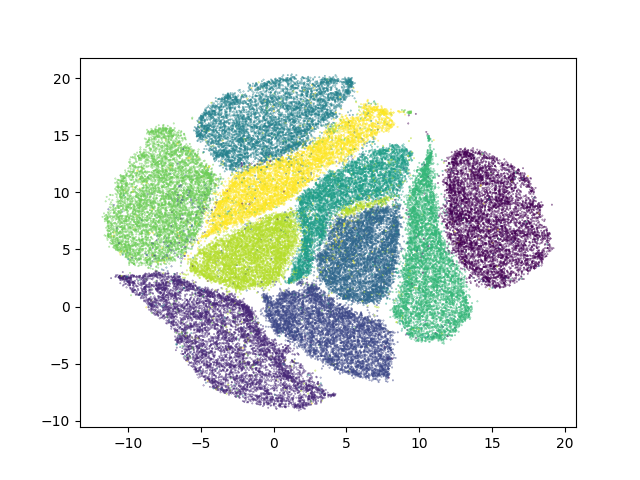}&
    \includegraphics[width=.123\linewidth]{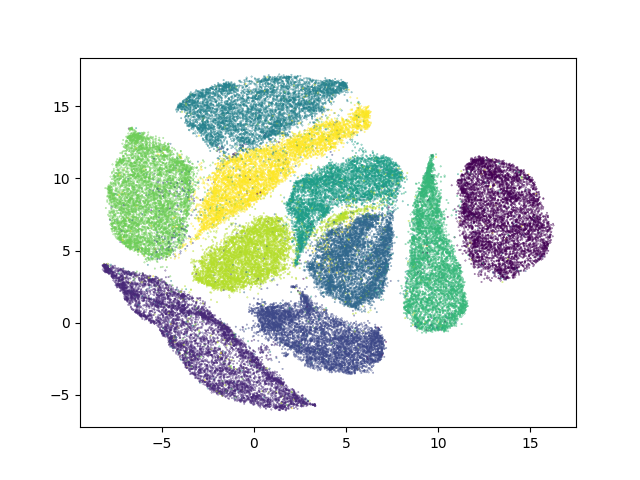}&
    \includegraphics[width=.123\linewidth]{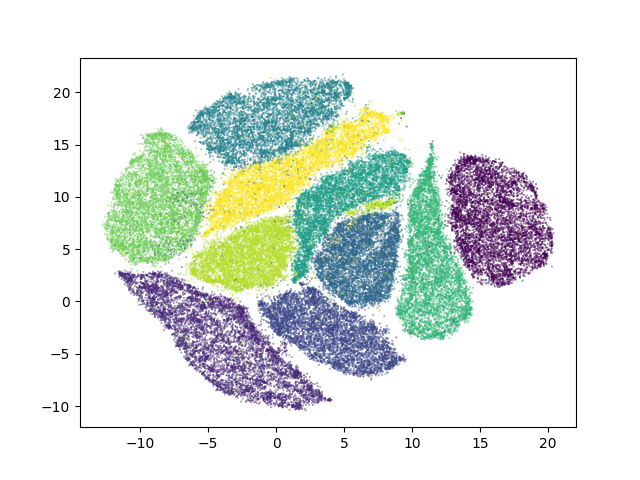}\\

    \includegraphics[width=.123\linewidth]{outputs/mnist/umap/default_embedding.png}&
    \includegraphics[width=.123\linewidth]{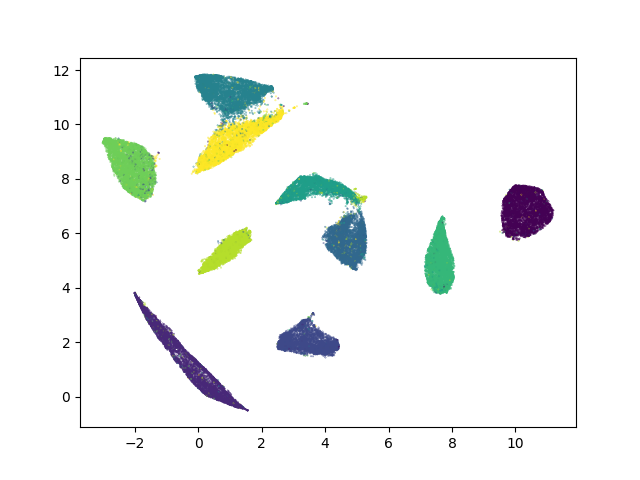}&
    \includegraphics[width=.123\linewidth]{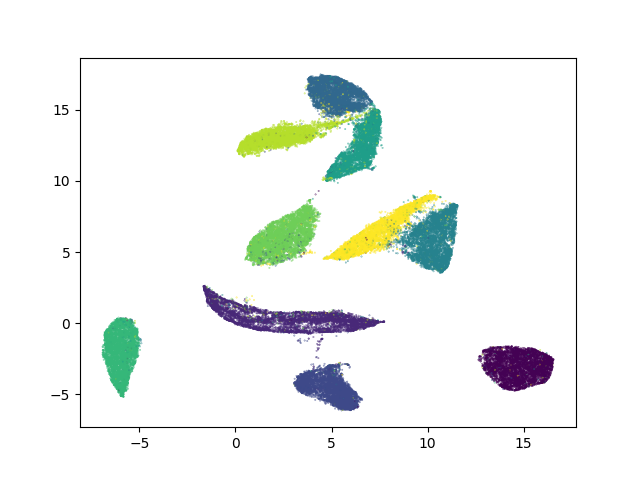}&
    \includegraphics[width=.123\linewidth]{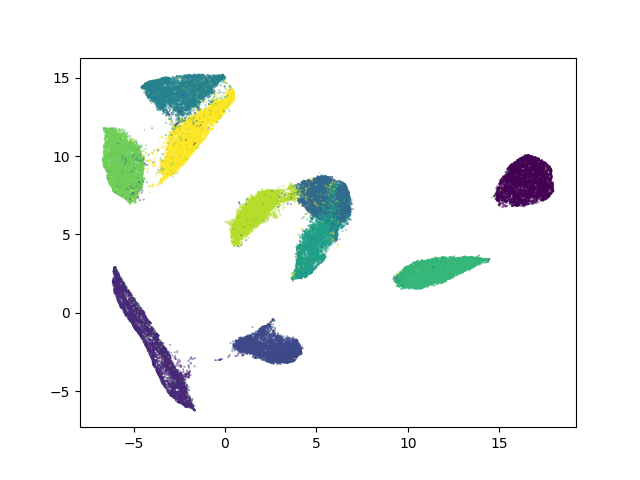}& 
    \includegraphics[width=.123\linewidth]{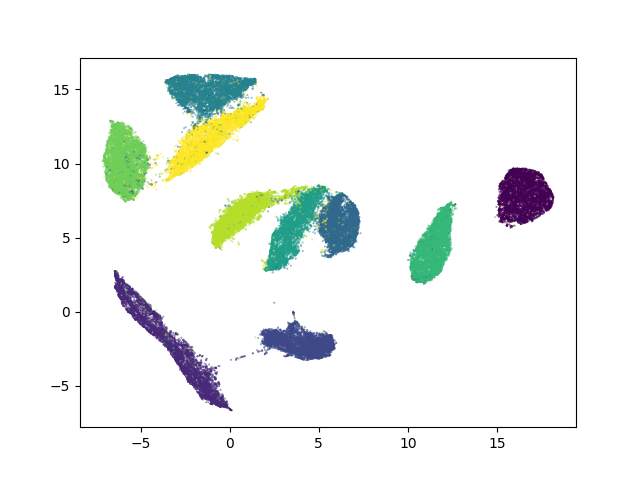}&
    \includegraphics[width=.123\linewidth]{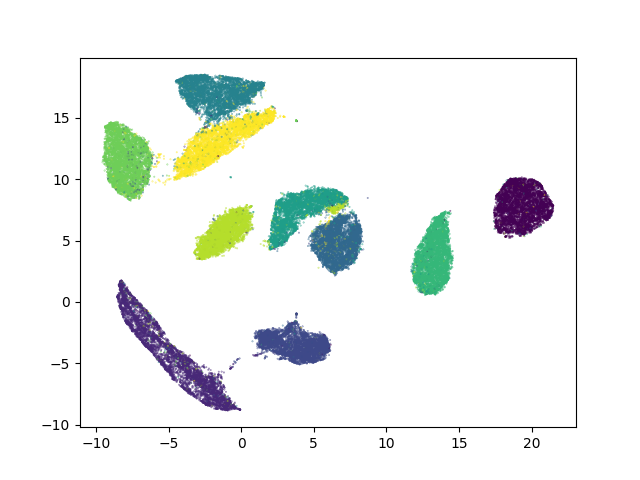}&
    \includegraphics[width=.123\linewidth]{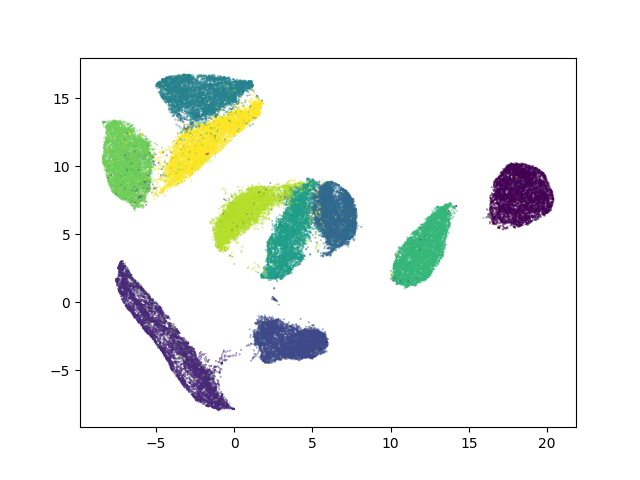}\\

    \includegraphics[width=.123\linewidth]{outputs/mnist/uniform_umap/default_embedding.png}&
    \includegraphics[width=.123\linewidth]{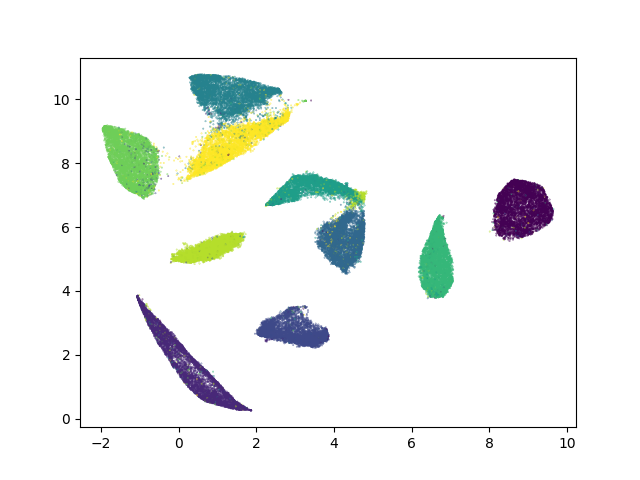}&
    \includegraphics[width=.123\linewidth]{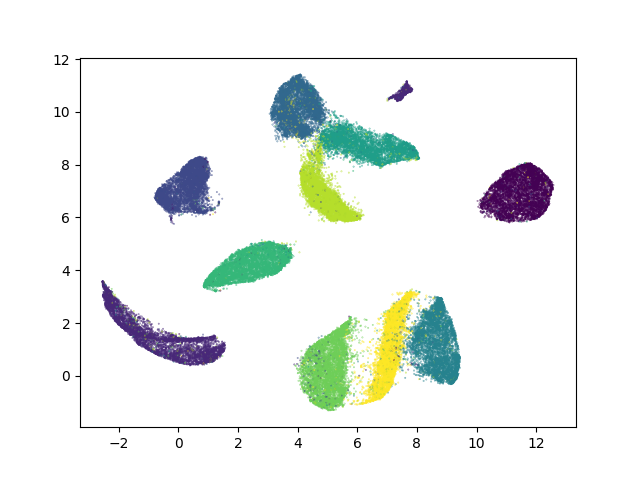}&
    \includegraphics[width=.123\linewidth]{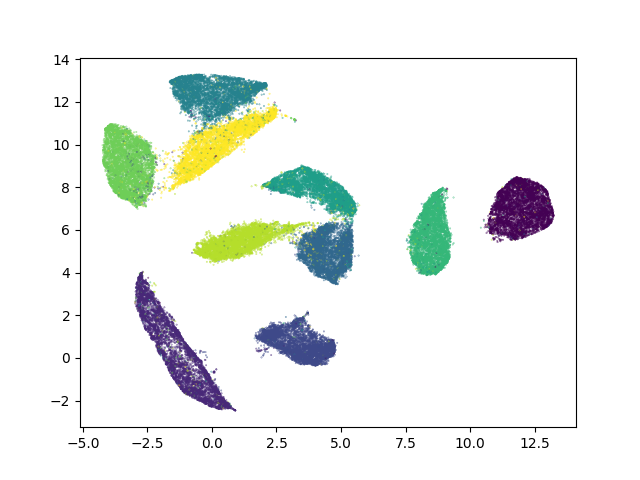}& 
    \includegraphics[width=.123\linewidth]{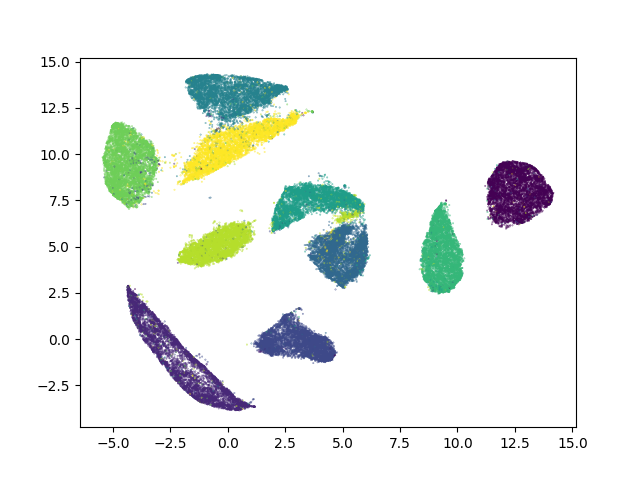}&
    \includegraphics[width=.123\linewidth]{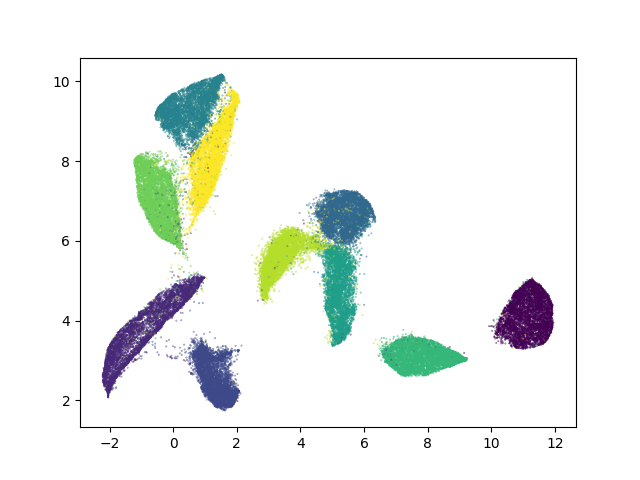}&
    \includegraphics[width=.123\linewidth]{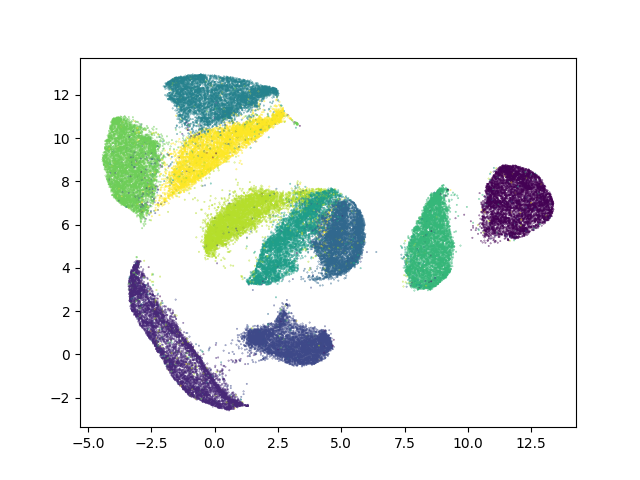}\\

    \bottomrule
    \end{tabularx}
    \caption{Effect of the algorithm settings from Table~\ref{differences_table} on MNIST. In each case, the parameter is changed from its default to its other option. For example, the initialization column initializes TSNE with Laplacian Eigenmaps, UMAP randomly and \ourmethod randomly -- the opposite of each algorithm's default. The rows are TSNE, \ourmethodN, UMAP, and \ourmethodU from top to bottom.}
    \label{irrelevant-mnist}
\end{table*}

\begin{table*}[!htb]
    \centering
    \setlength{\aboverulesep}{0pt}
    \setlength{\belowrulesep}{0pt}
    \begin{subtable}[t]{\linewidth}
        \begin{tabularx}{\linewidth}{m{1.5cm}|*{7}{c}|c}
         \textit{Algorithm} & \textit{Original} & Frobenius & Initialization & Pseudo-distance & Symmetrization & Sym attraction & Scalars & \textit{Algorithm mean} \\
        \midrule
        UMAP          & 95.4 & 96.7 & 96.6 & 94.4 & 96.7 & 96.6 & 96.5 & 96.3$\pm$1.0 \\ 
        \ourmethodU   & 96.2 & 96.2 & 96.4 & 96.7 & 96.6 & 96.5 & 95.8 & 96.3$\pm$0.3 \\ 
        TSNE          & 95.1 & 94.7 & 95.2 & 96.0 & 94.9 & 94.8 & 95.1 & 95.1$\pm$0.4\\ 
        \ourmethodN   & 96.1 & 96.3 & 95.6 & 96.1 & 96.1 & 96.3 & 96.1 & 96.1$\pm$0.2 \\ 
        \midrule
        \textit{Parameter mean dev.} & -0.3 & 0.1 & 0.0 & 0.1 & 0.4 & -0.7 &  0.4 & \\
        \end{tabularx}    
    \caption{kNN-accuracy}        
    \end{subtable}
    \begin{subtable}[t]{\textwidth}
        \begin{tabularx}{\linewidth}{m{1.5cm}|*{7}{c}|c}
         \textit{Algorithm} & \textit{Original} & Frobenius & Initialization & Pseudo-distance & Symmetrization & Sym attraction & Scalars & \textit{Algorithm mean} \\
        \midrule
        UMAP          & 82.5 & 87.0 & 84.6 & 82.2 & 82.5 & 83.5 & 82.2 & 83.5$\pm$1.6 \\ 
        \ourmethodU   & 84.0 & 85.4 & 82.1 & 85.2 & 85.1 & 83.3 & 81.2 & 83.8$\pm$1.5 \\ 
        TSNE          & 70.9 & 71.8 & 70.7 & 73.9 & 70.8 & 80.7 & 73.5 & 73.2$\pm$3.3 \\ 
        \ourmethodN   & 67.8 & 71.9 & 61.3 & 63.0 & 68.4 & 72.7 & 68.8 & 67.7$\pm$3.9 \\ 
        \midrule
        \textit{Parameter mean dev.} & -1.4 & 2.2 & -2.4 & -1.0 & 1.3 & -1.0 & 1.4 & \\
        \end{tabularx}
    \caption{V-score}    
    \end{subtable}
    \caption{kNN-accuracy ($k=100$) and V-score on MNIST for each parameter. Algorithm mean shows mean and confidence intervals on all the algorithmic changes. 
     \ourmethod successfully produces TSNE and UMAP embeddings across the settings. Parameter mean deviation shows the average difference between each value and the algorithm mean. A value close to 0 means no impact.}
    \label{irrelevant-metrics}
\end{table*}

\subsection{TSNE \& UMAP paramater study} 
\label{ssec:irrelevant_alg_params}

In this set of experiments, we show that the normalization is the main parameter responsible for the differences between TSNE and UMAP.  
Our first step is to show that the settings other than the normalization do not significantly affect the embeddings. In the first columns of Tables~\ref{irrelevant-mnist}, \ref{irrelevant-fashion-mnist} and Table~\ref{irrelevant-metrics}, we report the original settings for each algorithm as described in Table~\ref{differences_table}. Every other column, then, shows the result of switching that specific parameter to its swapped setting. For example, the initialization column implies that TSNE was initialized with Laplacian Eigenmaps while UMAP was initialized from a random distribution, since these are the \textit{opposites} of their default settings. Comparing to the default algorithms' embeddings, we see that none of the parameters impose an immediately visible effect on the resulting images. Table~\ref{irrelevant-metrics} shows that the parameter modifications do not significantly affect the kNN-accuracy or V-measure. The largest changes can be seen in the Frobenius norm, the initialization, and the symmetric attraction. We discuss each of these below.

\begin{table}
    \centering
    \begin{tabular}{|c|c|c|c|}
    \hline
    TSNE & \makecell{Our TSNE} & UMAP & \makecell{Our UMAP} \\
    \hline
    731.5 $\pm$ 22.3 & 430.3$\pm$42.5 & 71.2 $\pm$ 0.8 & 38.9$\pm$0.5 \\
    \hline
    \end{tabular}
    \caption{Runtimes (in seconds) for the original TSNE and UMAP algorithms vs. our implementations of TSNE and UMAP on the MNIST dataset.}
    \label{original_mnist_speeds}
\end{table}

\para{Frobenius Norm.} We observe stronger repulsions than the KL-divergence in the normalized setting, as seen in Table~\ref{grad_plots}. This imposes slightly more separation between dissimilar points in the low-dimensional space and makes for tighter clusters of similar points. Despite this qualitative change, the metrics across datasets are largely invariant to the change to Frobenius norm.

\para{Initialization.} Initialization naturally affects the layout of the points due to the highly non-convex objective function. While the relative positions of the clusters change based on the random initializations, the relevant intra-cluster distances remain consistent. Namely, clusters that are far apart can be placed anywhere in the embedding as long as they maintain sufficient distance from one another. This is a clear reflection of the large areas of $0$-magnitude gradients in every plot in Figure \ref{grad_plots}. In either case, there does not seem to be a discernible change in quality between the two initializations, leading us to prefer the Laplacian Eigenmap initialization on small datasets ($<100K$) due to its predictable output and random initialization on large datasets ($>100K$) due to computational concerns.

\para{Symmetric attraction.} If we enable symmetric attraction, $y_j$ is attracted to $y_i$ when we attract $y_i$ to $y_j$; otherwise, the attraction only gets applied onto $y_i$. Thus, switching from asymmetric to symmetric attraction is functionally similar to scaling the attractive force by 2. That leads to tighter clusters in the case of TSNE, whose clusters are generally spread out across the embedding. However, UMAP already tends to have tighter clusters than TSNE, so symmetric attraction plays a smaller role both qualitatively and quantitatively. We thus chose to implement \ourmethod with asymmetric attraction as it adequately recreates UMAP embeddings, better approximates the TSNE outputs, and is also quicker to optimize.

\smallskip 
We only show the effect of single hyperparameter changes for combinatorial reasons. In our experiments, however, we see no significant difference between applying one hyperparameter change or any other number of them. Furthermore, there are a few hyperparameters that we do not include in the main body of this paper, as they both have no effect on the embeddings and are the least interesting. These include exact vs. approximate nearest neighbor calculation, gradient clipping, and the number of epochs.

In order to evaluate each of the above changes, we re-implemented the TSNE and UMAP algorithms such that each individual setting can be easily enabled or disabled. There are a few incompatibilities that we would like to mention, however. The fact that UMAP applies gradients inside the optimization loop is incompatible with both momentum gradient descent and the change in normalization. This is due to the fact that both of these rely on scalars that are gathered over the course of the entire epoch, and thus are unavailable to us during the optimization loop. As such, both the normalization and gradient amplification were evaluated on a variant of UMAP in which gradients are collected across the epoch and then applied. Although this means that we are not completely isolating the settings in question, we find that collecting gradients vs. applying them immediately does not change the resulting embedding in any of our experiments, giving us sufficient confidence that the analysis is still relevant.


\begin{table*}
    \newcolumntype{C}{ >{\centering\arraybackslash} m{2.3cm} }
    \newcolumntype{D}{ >{\centering\arraybackslash} m{1.5cm} }
    \centering
    \begin{tabular}{DCCCCCC}
    \multicolumn{1}{c}{} & \multicolumn{2}{c}{MNIST} & \multicolumn{2}{c}{Fashion-MNIST} & \multicolumn{2}{c}{Swiss Roll} \\
    \cmidrule(lr){2-3}\cmidrule(lr){4-5}\cmidrule(lr){6-7}
    \multicolumn{1}{c}{} & \multicolumn{1}{c}{\makecell{TSNE \\ Normalization}} & \multicolumn{1}{c}{\makecell{UMAP \\ Normalization}}
    & \multicolumn{1}{c}{\makecell{TSNE \\ Normalization}} & \multicolumn{1}{c}{\makecell{UMAP \\ Normalization}}
    & \multicolumn{1}{c}{\makecell{TSNE \\ Normalization}} & \multicolumn{1}{c}{\makecell{UMAP \\ Normalization}}\\
    
    \makecell{TSNE} &
    \includegraphics[width=.95\linewidth]{outputs/mnist/tsne/default_embedding.png}&
    \includegraphics[width=.95\linewidth]{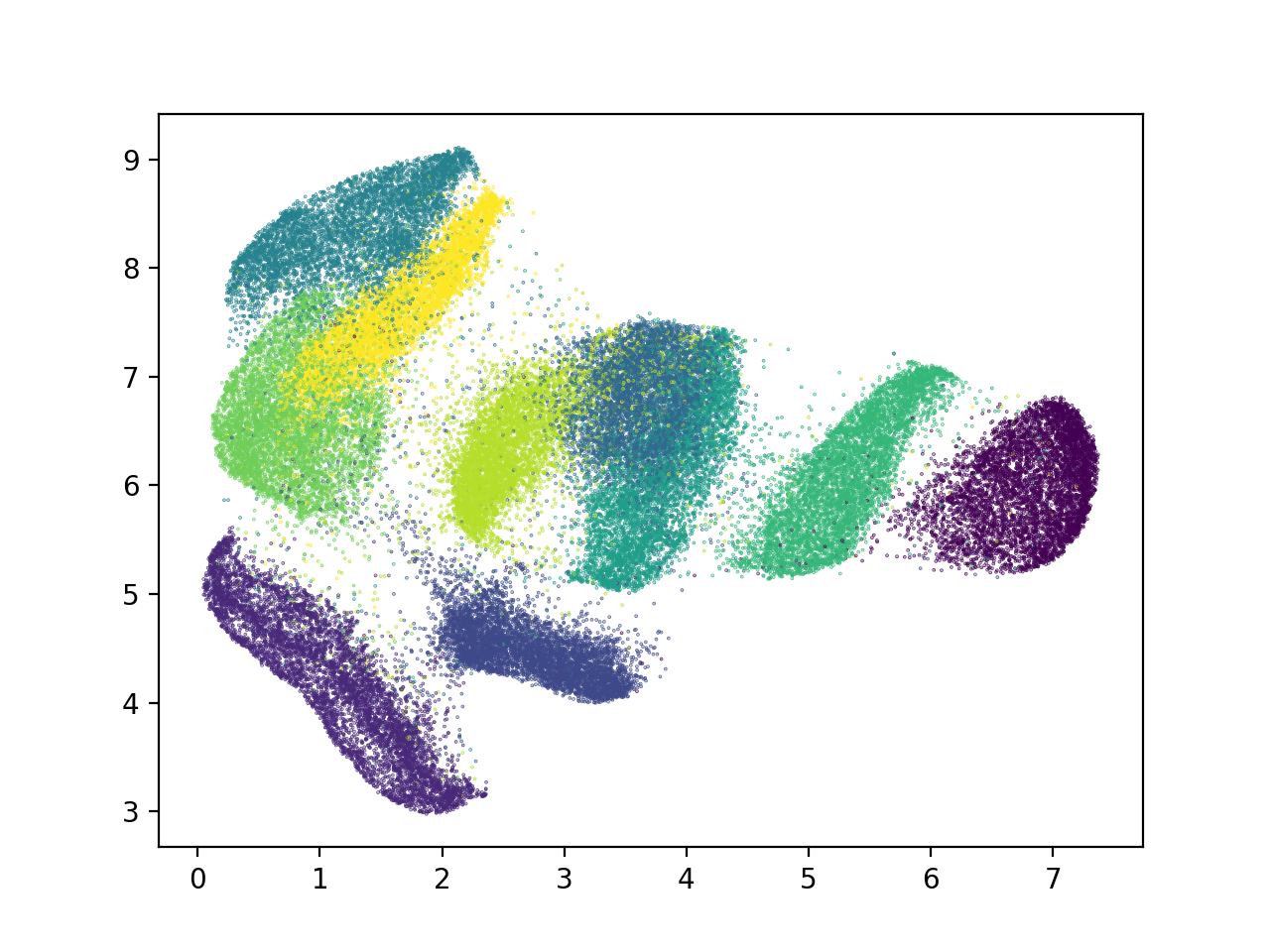}&
    \includegraphics[width=.95\linewidth]{outputs/fashion_mnist/tsne/default_embedding.png}&
    \includegraphics[width=.95\linewidth]{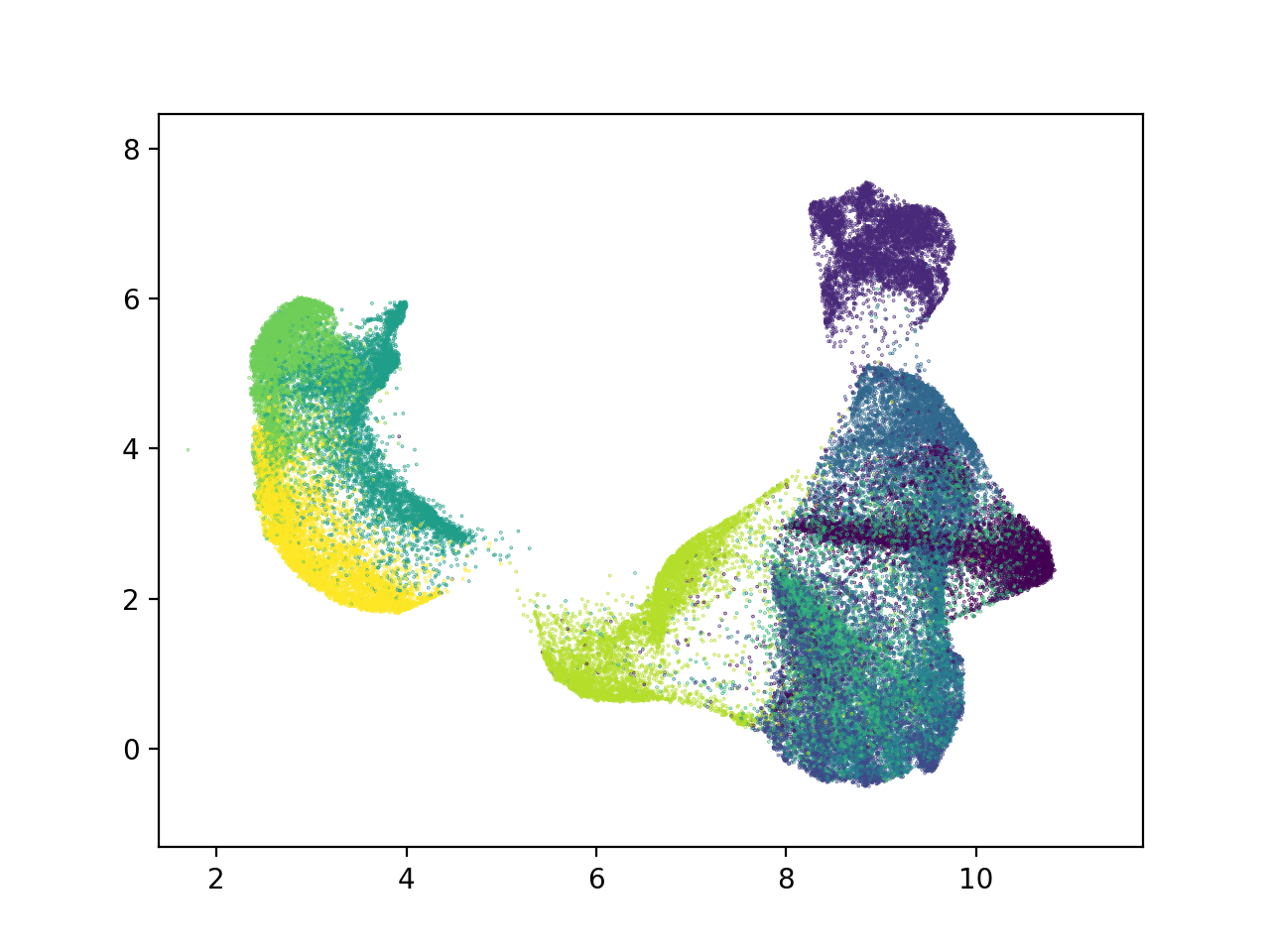}&
    \includegraphics[width=.95\linewidth]{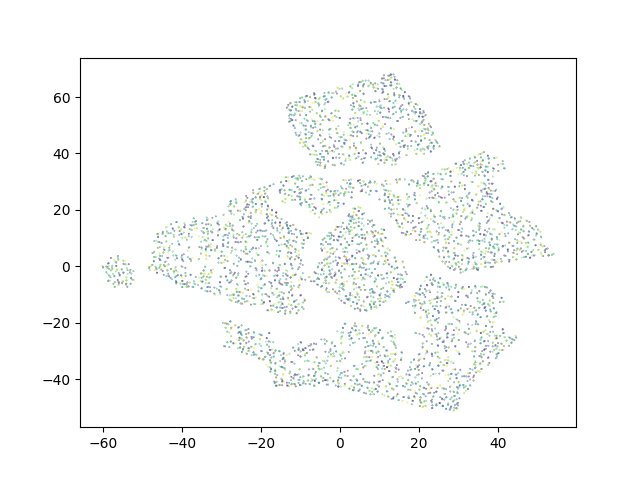}&
    \includegraphics[width=.95\linewidth]{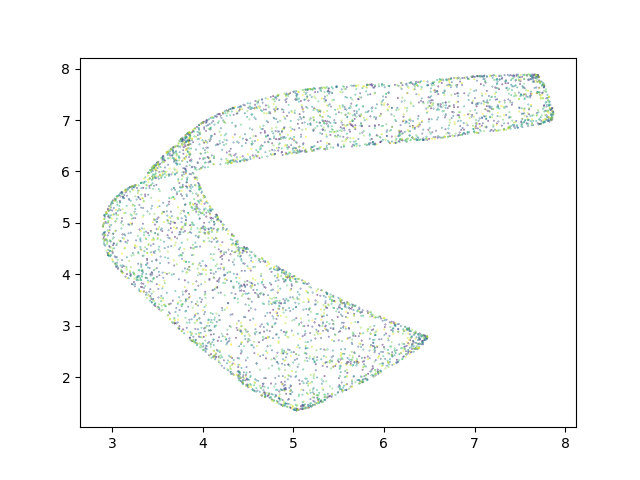}\\
    
    \makecell{UMAP} &
    \includegraphics[width=.95\linewidth]{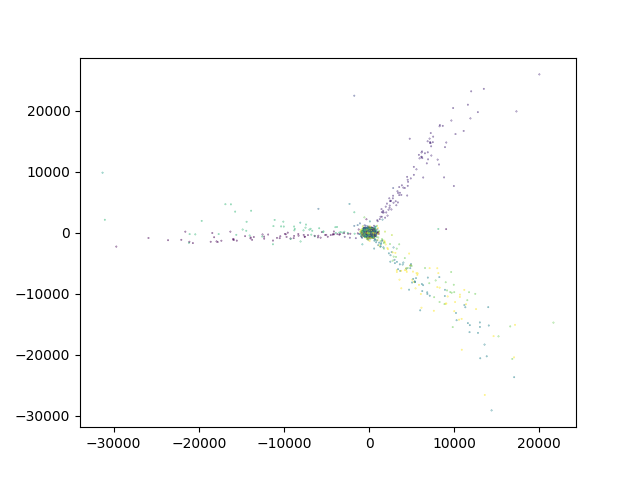}&
    \includegraphics[width=.95\linewidth]{outputs/mnist/umap/default_embedding.png}&
    \includegraphics[width=.95\linewidth]{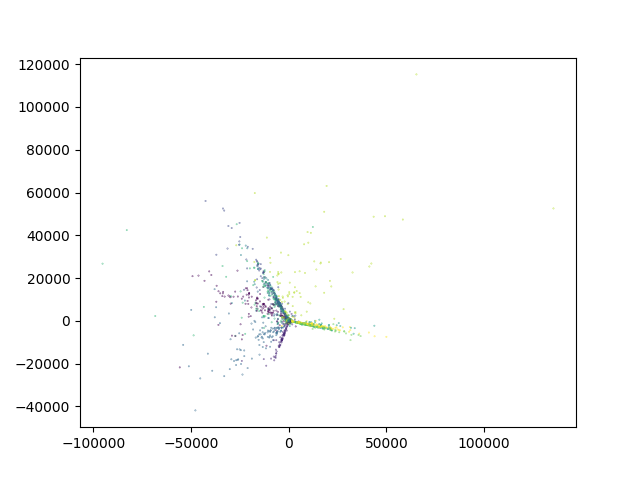}&
    \includegraphics[width=.95\linewidth]{outputs/fashion_mnist/umap/default_embedding.png}&
    \includegraphics[width=.95\linewidth]{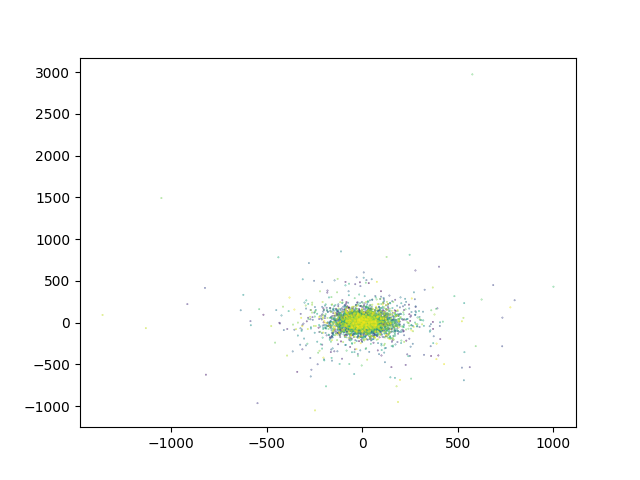}&
    \includegraphics[width=.95\linewidth]{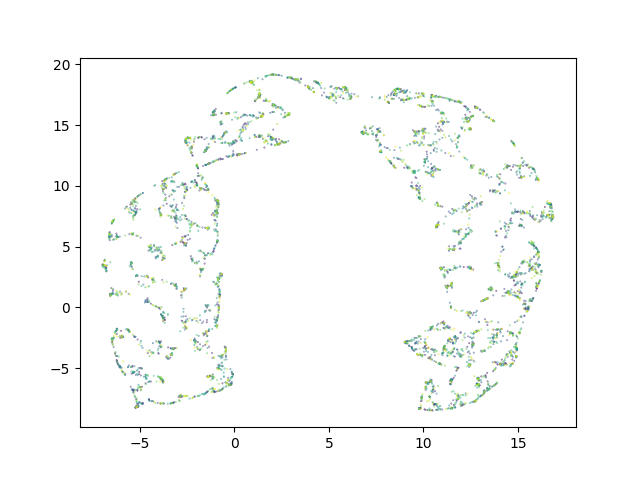}
    \\
    
    \makecell{\ourmethod} &
    \includegraphics[width=.95\linewidth]{outputs/mnist/uniform_umap/normalized_embedding.png}&
    \includegraphics[width=.95\linewidth]{outputs/mnist/uniform_umap/default_embedding.png}&
    
    \includegraphics[width=.95\linewidth]{outputs/fashion_mnist/uniform_umap/normalized_embedding.png}&
    \includegraphics[width=.95\linewidth]{outputs/fashion_mnist/uniform_umap/default_embedding.png}&
    
    \includegraphics[width=.95\linewidth]{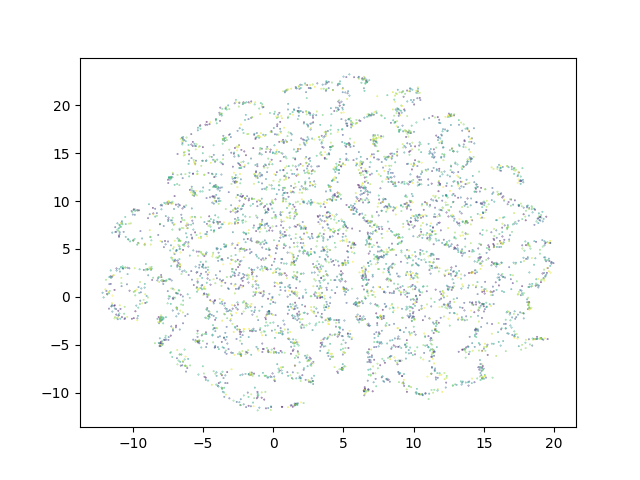}&
    \includegraphics[width=.95\linewidth]{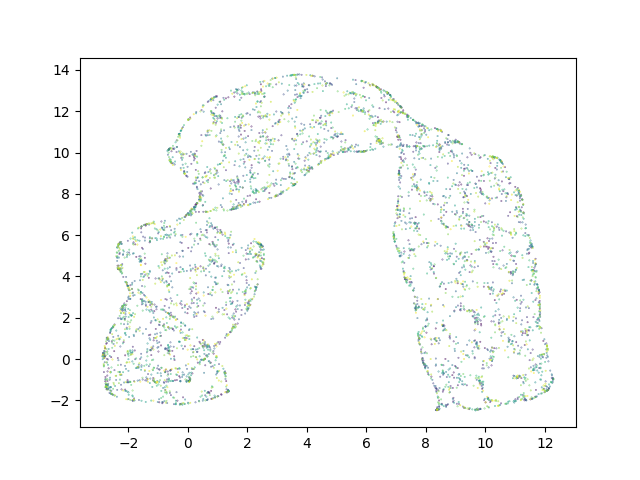}
    \\
    \end{tabular}
    \caption{Effect of changing the normalization on the MNIST, Fashion-MNIST, and Swiss Roll datasets. Notice that \ourmethod successfully recreates both the TSNE and UMAP results. We use Laplacian Eigenmap initializations for consistent orientation.}
    \label{relevant-mnist}
\end{table*}

\begin{table*}
    \centering
    \begin{tabularx}{\textwidth}{c*{6}{c}}
    & \multicolumn{6}{c}{\textit{Swapped setting}}\\
    \cline{2-7}
    \textit{Original setting} & Frobenius & Initialization & Pseudo-distance & Symmetrization & Sym attraction & Scalars\\

    \midrule

    \includegraphics[width=.123\linewidth]{outputs/fashion_mnist/tsne/default_embedding.png}&
    \includegraphics[width=.123\linewidth]{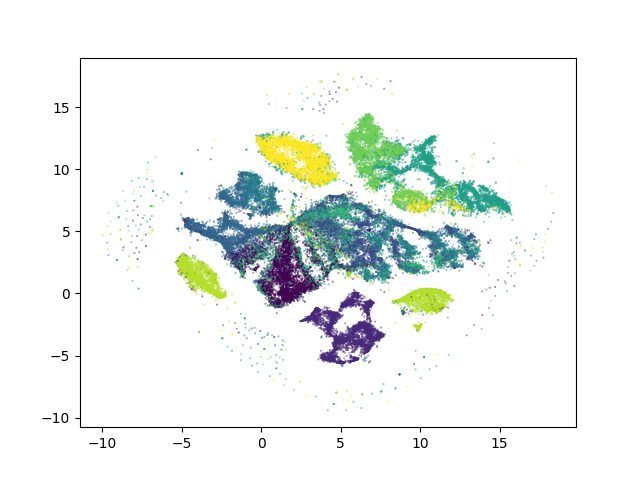}&
    \includegraphics[width=.123\linewidth]{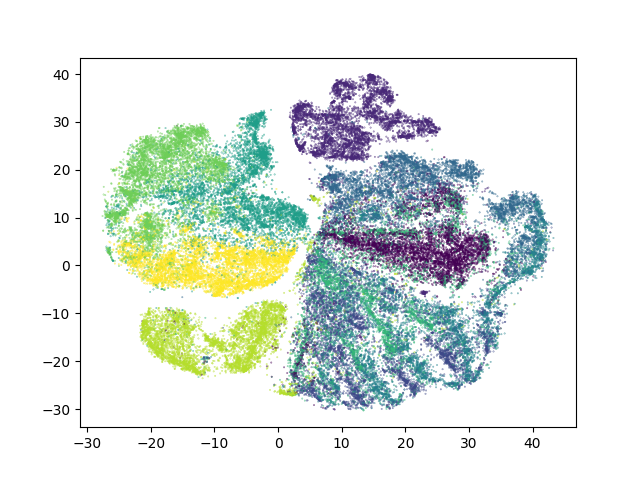}&
    \includegraphics[width=.123\linewidth]{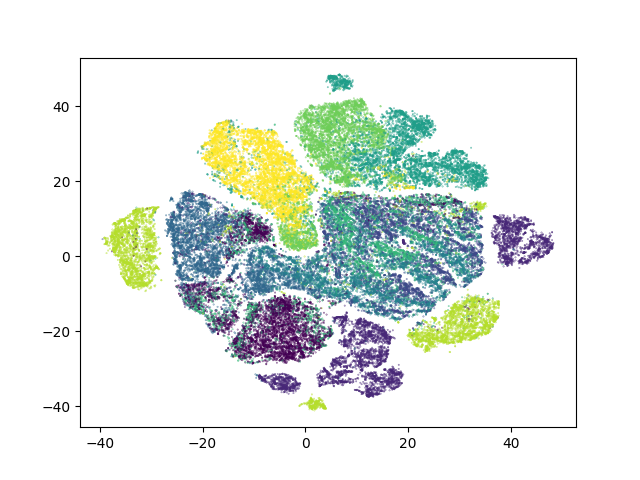}& 
    \includegraphics[width=.123\linewidth]{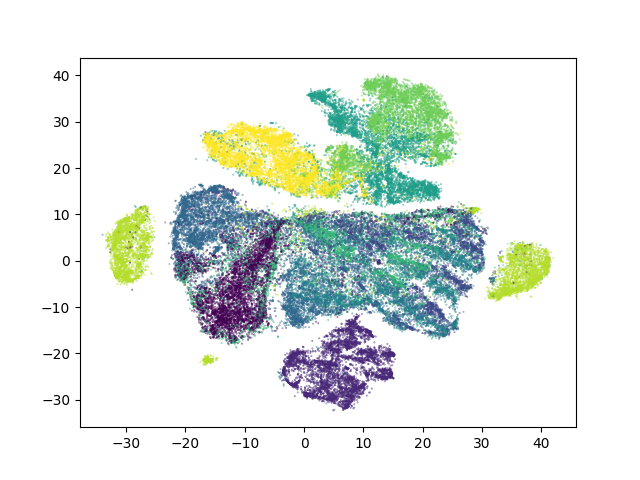}&
    \includegraphics[width=.123\linewidth]{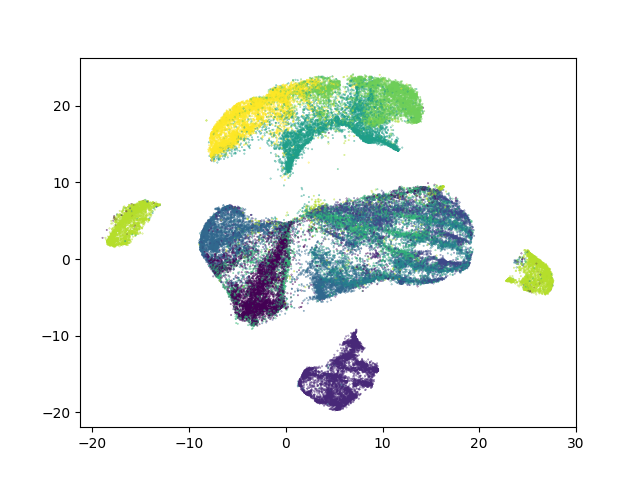}&
    \includegraphics[width=.123\linewidth]{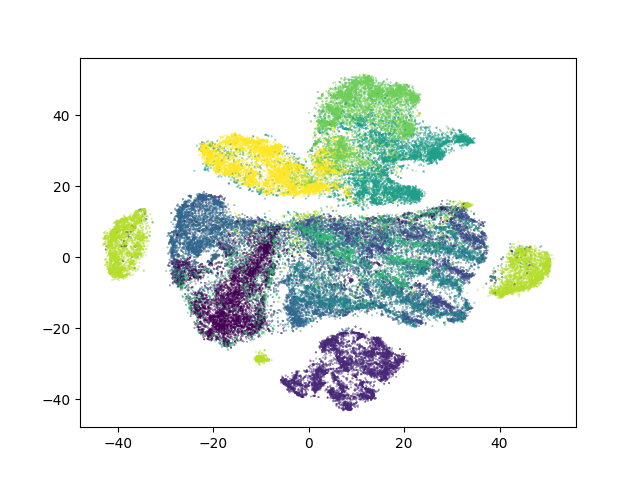}\\
    
    \includegraphics[width=.123\linewidth]{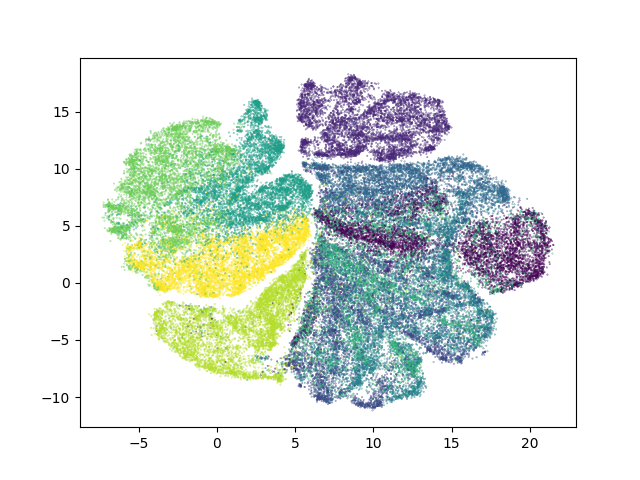}&
    \includegraphics[width=.123\linewidth]{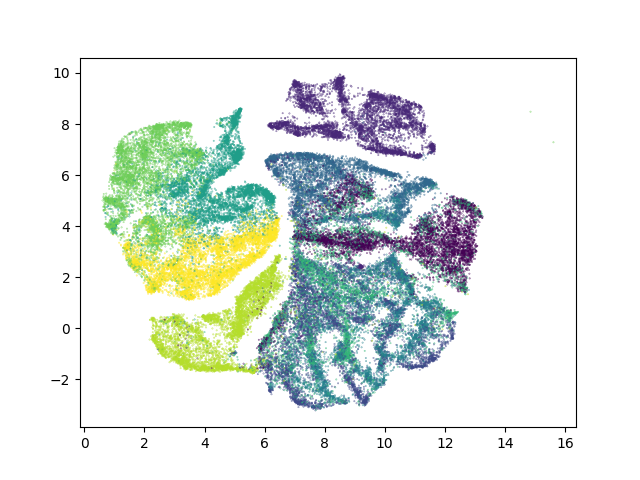}&
    \includegraphics[width=.123\linewidth]{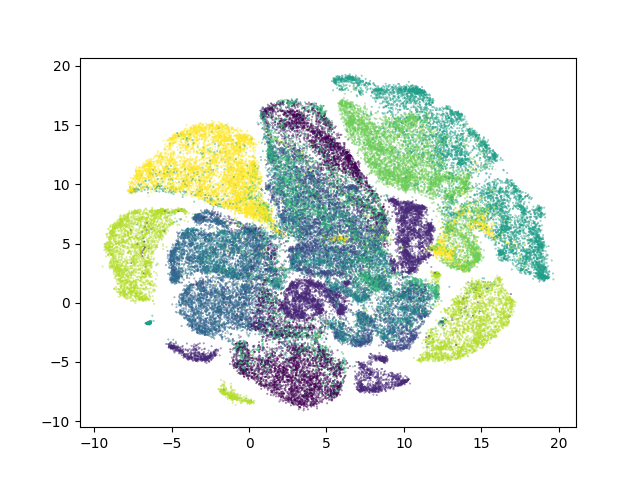}&
    \includegraphics[width=.123\linewidth]{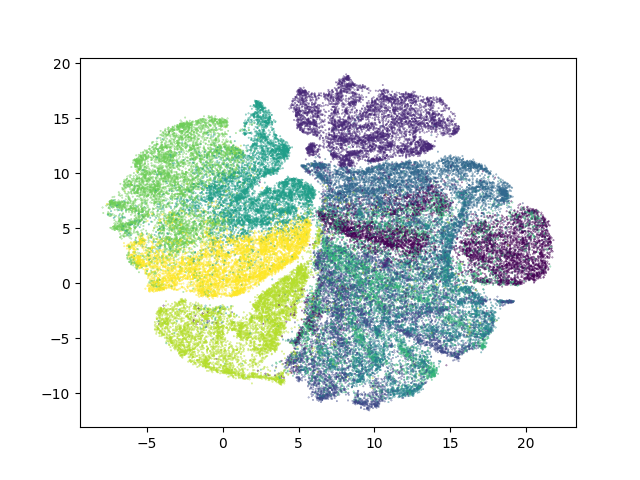}& 
    \includegraphics[width=.123\linewidth]{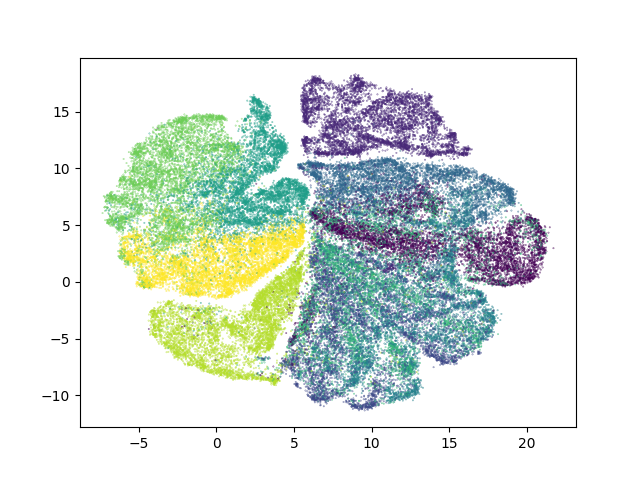}&
    \includegraphics[width=.123\linewidth]{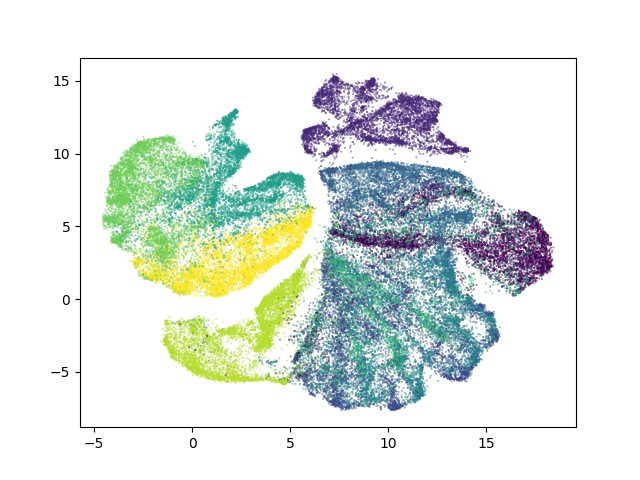}&
    \includegraphics[width=.123\linewidth]{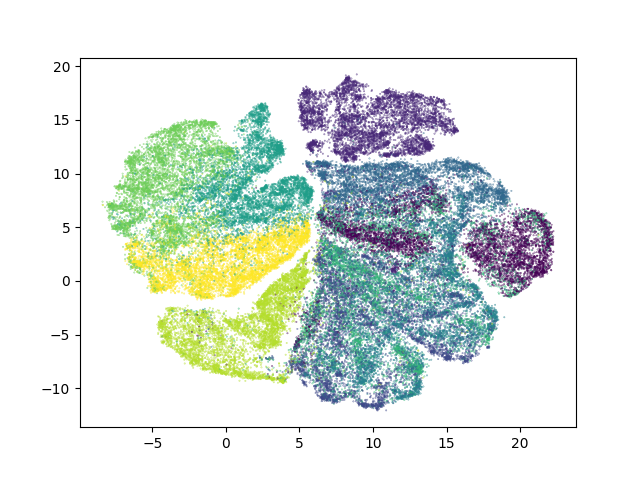}\\

    \includegraphics[width=.123\linewidth]{outputs/fashion_mnist/umap/default_embedding.png}&
    \includegraphics[width=.123\linewidth]{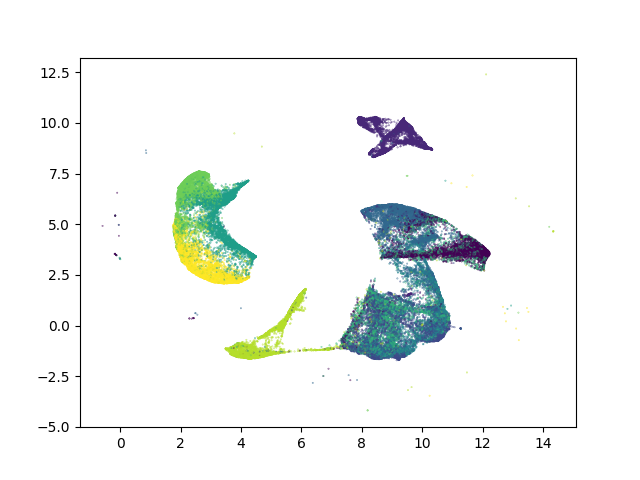}&
    \includegraphics[width=.123\linewidth]{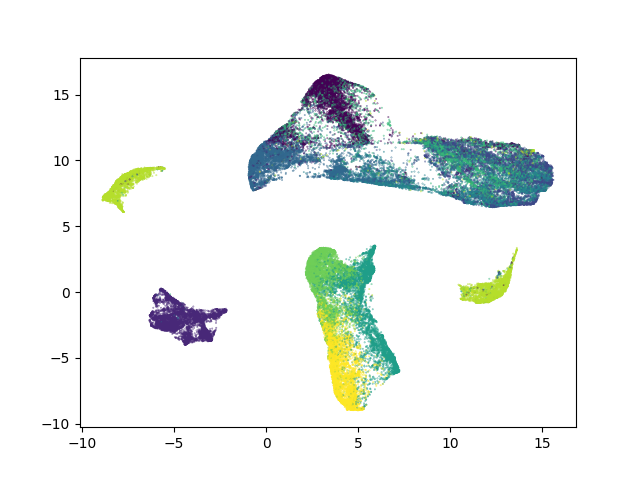}&
    \includegraphics[width=.123\linewidth]{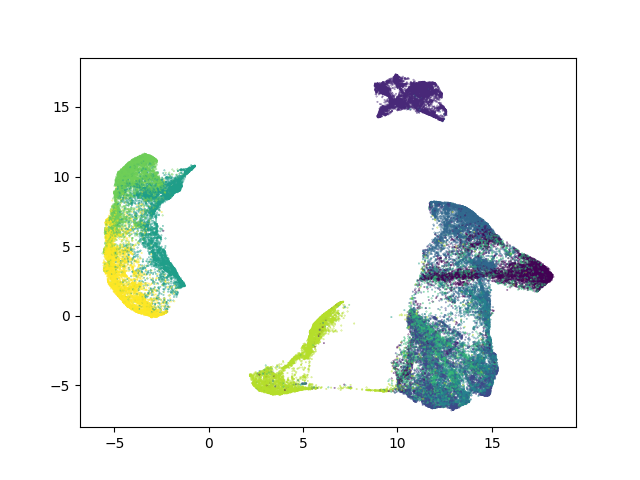}& 
    \includegraphics[width=.123\linewidth]{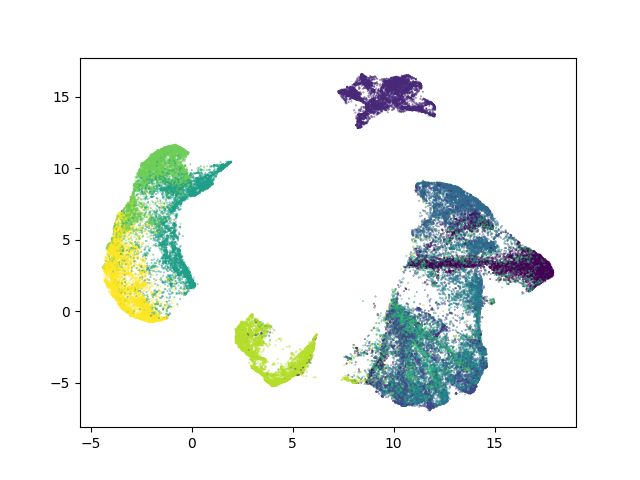}&
    \includegraphics[width=.123\linewidth]{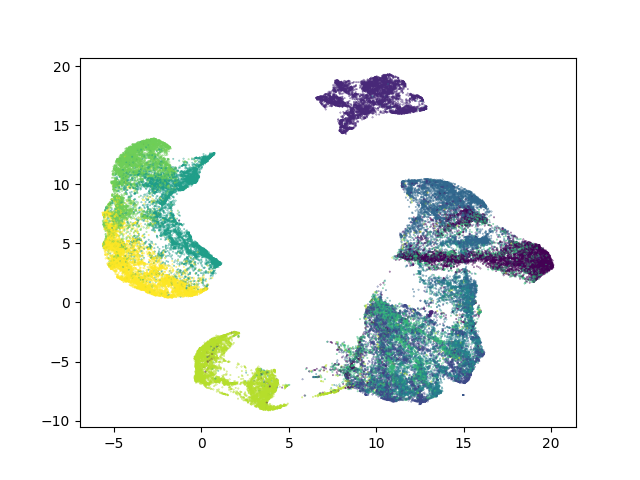}&
    \includegraphics[width=.123\linewidth]{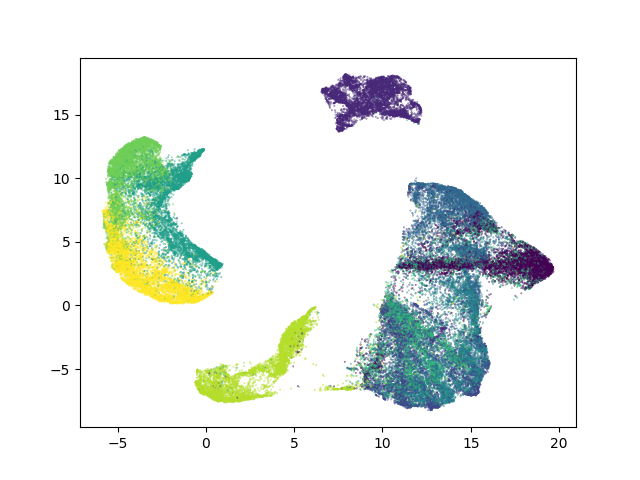}\\

    \includegraphics[width=.123\linewidth]{outputs/fashion_mnist/uniform_umap/default_embedding.png}&
    \includegraphics[width=.123\linewidth]{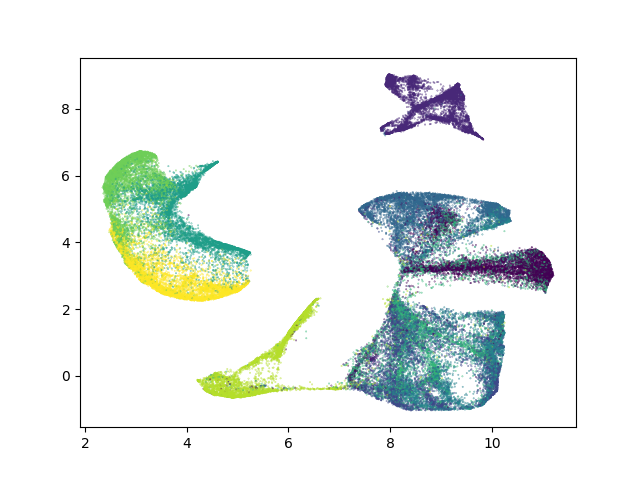}&
    \includegraphics[width=.123\linewidth]{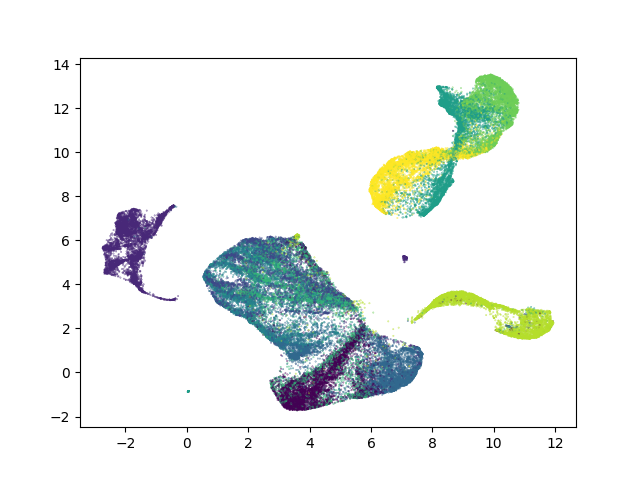}&
    \includegraphics[width=.123\linewidth]{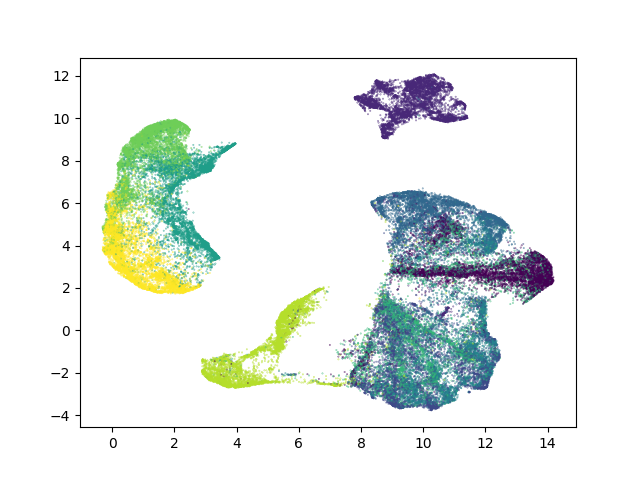}& 
    \includegraphics[width=.123\linewidth]{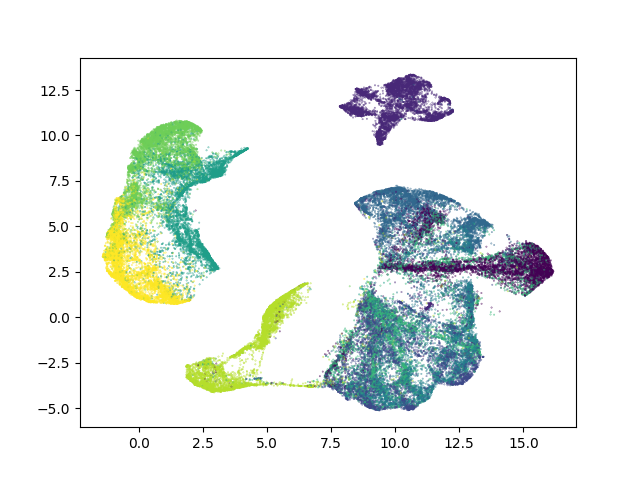}&
    \includegraphics[width=.123\linewidth]{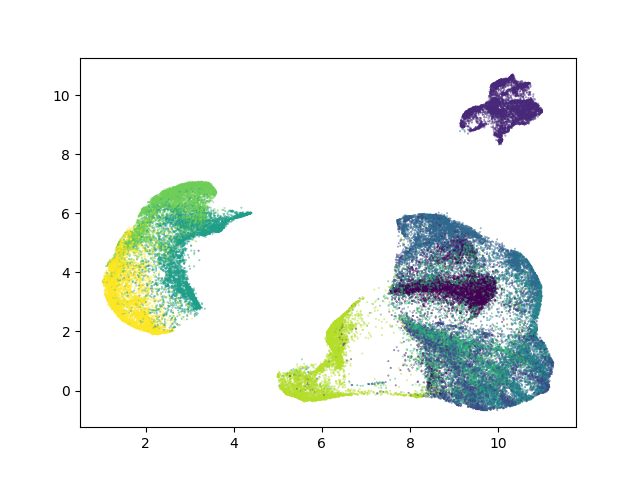}&
    \includegraphics[width=.123\linewidth]{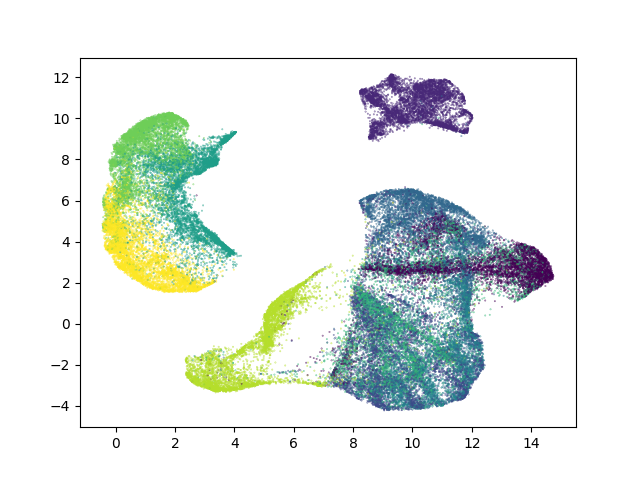}\\

    \bottomrule
    \end{tabularx}
    \caption{Effect of the algorithm settings from Table \ref{differences_table} on Fashion MNIST embeddings. The rows are TSNE, \ourmethodN, UMAP, and \ourmethodU from top to bottom.}
    \label{irrelevant-fashion-mnist}
\end{table*}


\begin{table}
\setlength{\aboverulesep}{0pt}
\setlength{\belowrulesep}{0pt}
    \newcolumntype{C}{>{\centering\arraybackslash}X}
    \centering
    \begin{tabularx}{\columnwidth}{C|C|CCC}
    \multicolumn{1}{C}{} & \multicolumn{1}{C}{\textit{Algorithm}} & \makecell{Fashion \\ MNIST} & Coil-100 & Cifar-10\\
    \midrule

    \multirow{4}{*}{\bf \makecell{kNN \\ Accuracy}} 
    & UMAP & 78.0$\pm$0.5 & 80.8$\pm$3.3 & 24.2$\pm$1.1 \\ 
    & \ourmethodU & 77.3$\pm$0.7 & 77.4$\pm$3.4 & 23.8$\pm$1.1 \\ 
    \cline{2-5}
    & TSNE & 80.1$\pm$0.7 & 63.2$\pm$4.2 & 28.7$\pm$2.5 \\ 
    & \ourmethodN & 78.6$\pm$0.6 & 77.2$\pm$ 4.4 & 25.6$\pm$1.1 \\
    \bottomrule
        
    \multirow{4}{*}{\bf V-score} 
    & UMAP & 60.3$\pm$1.4 & 89.2$\pm$0.9 & 7.6$\pm$0.4\\ 
    & \ourmethodU & 61.7$\pm$0.8 & 91.0$\pm$0.6 & 8.1$\pm$0.6 \\
    \cline{2-5}
    & TSNE & 54.2$\pm$4.1 & 82.9$\pm$1.8 & 8.5$\pm$0.3 \\ 
    & \ourmethodN & 51.7$\pm$4.7 & 85.7$\pm$2.6 & 8.0$\pm$3.7 \\
    \bottomrule
    
    \end{tabularx}
    \caption{Algorithm mean of kNN-accuracy and V-score as in Table~\ref{irrelevant-metrics} for Fashion MNIST and Coil-100. \ourmethod replicates UMAP and TSNE results across datasets.}
    \label{irrelevant-metrics-other-data}
\end{table}


        
    

\begin{table*}
    \setlength{\aboverulesep}{0pt}
    \setlength{\belowrulesep}{0pt}
    \newcolumntype{C}{>{\centering\arraybackslash}X}
    \centering
    \begin{tabularx}{\linewidth}{C|c|*{7}{c}}
     & \textit{Dataset} & \textit{Original} & Frobenius & Init. & Pseudo-distance & Symmetrization & Sym attraction & Scalars\\
    \midrule

    \multirow{3}{*}{kNN-accuracy}
    & Fashion-MNIST     & -0.7  & 1.1   & -0.4 & -0.3   & 0.4 & -0.8     & 0.7   \\ 
    & Coil-100          & 0.8   & -6.9  & -2.8 & 3.7    & 2.3 & 1.1      & 1.8   \\ 
    & Cifar-10          & -0.3  & 2.2   & -1.0 & -0.2   & 0.8 & -1.3     & -0.1  \\
    \midrule
    \multirow{3}{*}{V-score}
    & Fashion-MNIST     & -1.3  & 2.5   & 1.5   & -0.8  & -1.6  & -0.8  & 0.4 \\ 
    & Coil-100          & 0.0   & -1.9  & -0.2  & 1.0   & 0.9   & -0.5  & 0.7 \\ 
    & Cifar-10          & -0.3  & 1.4   & -0.3  & -0.3  & 0.1   & -0.4  & -0.1 \\
    \bottomrule
    
    \end{tabularx}
    \caption{Parameter mean deviation as in Table~\ref{irrelevant-metrics} for Fashion-MNIST, Coil-100 and Cifar-10. A value close to 0 implies that the parameter does not affect the embeddings.}
    \label{irrelevant-metrics_col_means}
\end{table*}

\subsection{Effect of Normalization} 
\label{ssec:norm_results}

We now analyze the effect that the normalization has on the embeddings. Across experiments, TSNE in the unnormalized setting induces significantly more separation between the clusters in a manner similar to UMAP. The representations are fuzzier than the UMAP ones, however, as we are still performing repulsions with respect to every other point, causing the embedding to fall closer to the mean of the multi-modal distribution. We draw the reader's attention to the fact that TSNE without normalization creates \dm{more} interpretable embeddings on the Fashion-MNIST and Swiss Roll datasets, as seen in Table~\ref{relevant-mnist}.

We see a much more extreme effect when performing UMAP in the normalized setting. Recall that the UMAP algorithm approximates the $p_{ij}$ and $1 - p_{ik}$ gradient scalars by sampling the attractions and repulsions proportionally to $p_{ij}$ and $1 - p_{ij}$, which we referred to as \textit{scalar sampling}. However, the gradients in the normalized setting (as seen in Equation~\ref{tsne_grad_equations}) still have the $p_{ij}$ scalar on attractions but lose the $1 - p_{ik}$ scalar on repulsions. The UMAP optimization schema, then, imposes an unnecessary weight on the repulsions in the normalized setting as the repulsions are still sampled according to the no-longer-necessary $1 - p_{ik}$ scalar. Accounting for this requires dividing the repulsive forces by $1 - p_{ik}$, but this (with the momentum gradient descent and stronger learning rate) leads to a highly unstable training regime, as can be seen by the normalized UMAP cells of Table~\ref{relevant-mnist}.

This implies that stabilizing UMAP in the normalized setting requires removing the sampling and instead directly multiplying by $p_{ij}$ and $1 - p_{ik}$. Indeed, this is exactly what we do in \ourmethod. Under this change and only this change, \ourmethodU obtains embeddings that are effectively identical to those that UMAP returns under its default parameters. Additionally, this is a simple paradigm to parallelize as the resulting blocks are consistent in complexity.

\subsection{Theoretical Implications}
Here we discuss how the above results fit into UMAP's theoretical framework. Much of UMAP's foundation is developed to justify the existence of a locally-connected manifold in the high-dimensional space under the UMAP pseudo-distance metric $\tilde{d}$. Recall that this pseudo-distance metric is defined such that the distance from point $x_j$ to $x_i$ is equal to $\tilde{d}(x_i, x_j) = d(x_i, x_j) - min_{k \neq i} d(x_i, x_k)$. This is clearly no longer a true distance metric and thus requires significant work before it can accommodate the locally-connected manifold arguments that UMAP is built on. Despite this clean foundation, however, our experimental results in Tables~\ref{irrelevant-mnist},~\ref{irrelevant-fashion-mnist}, and~\ref{irrelevant-metrics} show that the embedding quality is not contingent on the pseudo-distance metric. We see three options for what this could mean. The first is that the datasets we have
chosen are not ones for which the pseudo-distance metric is necessary. This seems unlikely, however, as we are operating across data domains and dimensionalities. The second option is that the algorithm's reliance on gradient descent and highly non-convex criteria deviates enough from the theoretical discussion that the pseudo-distance metric loses its applicability. The last option is that this pseudo-distance metric, while insightful from a theoretical perspective, is not a necessary calculation in order to achieve the final embeddings.

We further investigate which algorithmic changes are compatible with the theoretical framework. For example, UMAP chose the KL divergence as a natural method for minimizing the difference between probability distributions. However, this choice was made arbitrarily and we see no reason that the Frobenius norm cannot be substituted for the KL divergence. Interestingly, Table~\ref{grad_plots} shows that optimizing the Frobenius norm gives significantly different attractive and repulsive forces. This adds nuance to the claim that UMAP is finding the single embedding that reproduces the high-dimensional manifold, as the embeddings for the Frobenius norm and KL divergence both satisfy the theoretical criteria despite having very different gradient descent processes. Furthermore, Table~\ref{grad_plots} shows that the zero-gradient areas between the KL divergence and the Frobenius norm strongly overlap, implying that a local minimum under one loss satisfies the other loss as well.

Additionally, we raise the question of whether the choice of normalization can fit into UMAP's theory. If the normalization does not break the assumptions in~\cite[Sec.~2,3]{mcinnes2018umap} of the original UMAP paper, we posit that the interpretation of UMAP as finding the best fit to the high-dimensional data manifold extends to TSNE under our formulation as well, as long as TSNE's gradients are calculated under the pseudo-distance metric in the high-dimensional space.

\subsection{\ourmethod vs. TSNE/UMAP} 
\label{ssec:alg_comparison_results}
Given the analysis of the TSNE and UMAP parameters, we compare the original algorithms with \ourmethod's outputs. Namely, we aim to show that the outputs of \ourmethodU and \ourmethodN are functionally equivalent to UMAP's and TSNE's respectively. This can be seen qualitatively in Table~\ref{irrelevant-mnist}, where we see that \ourmethod reproduces TSNE and UMAP embeddings across hyperparameters. This is similarly evident in the metrics, where \ourmethodU replicates UMAP and \ourmethodN replicates TSNE for both kNN accuracy and K-means V-score in Tables~\ref{irrelevant-metrics} and~\ref{irrelevant-metrics-other-data}. We cannot compare \ourmethod to TSNE and UMAP under differing normalizations, as that is the setting dictating which algorithm \ourmethod is emulating. Thus, across all of the identified hyperparameters and settings, our method satisfactorily reproduces the embeddings for both TSNE and UMAP.

\begin{figure}[!htb]
    \includegraphics[width=.95\linewidth]{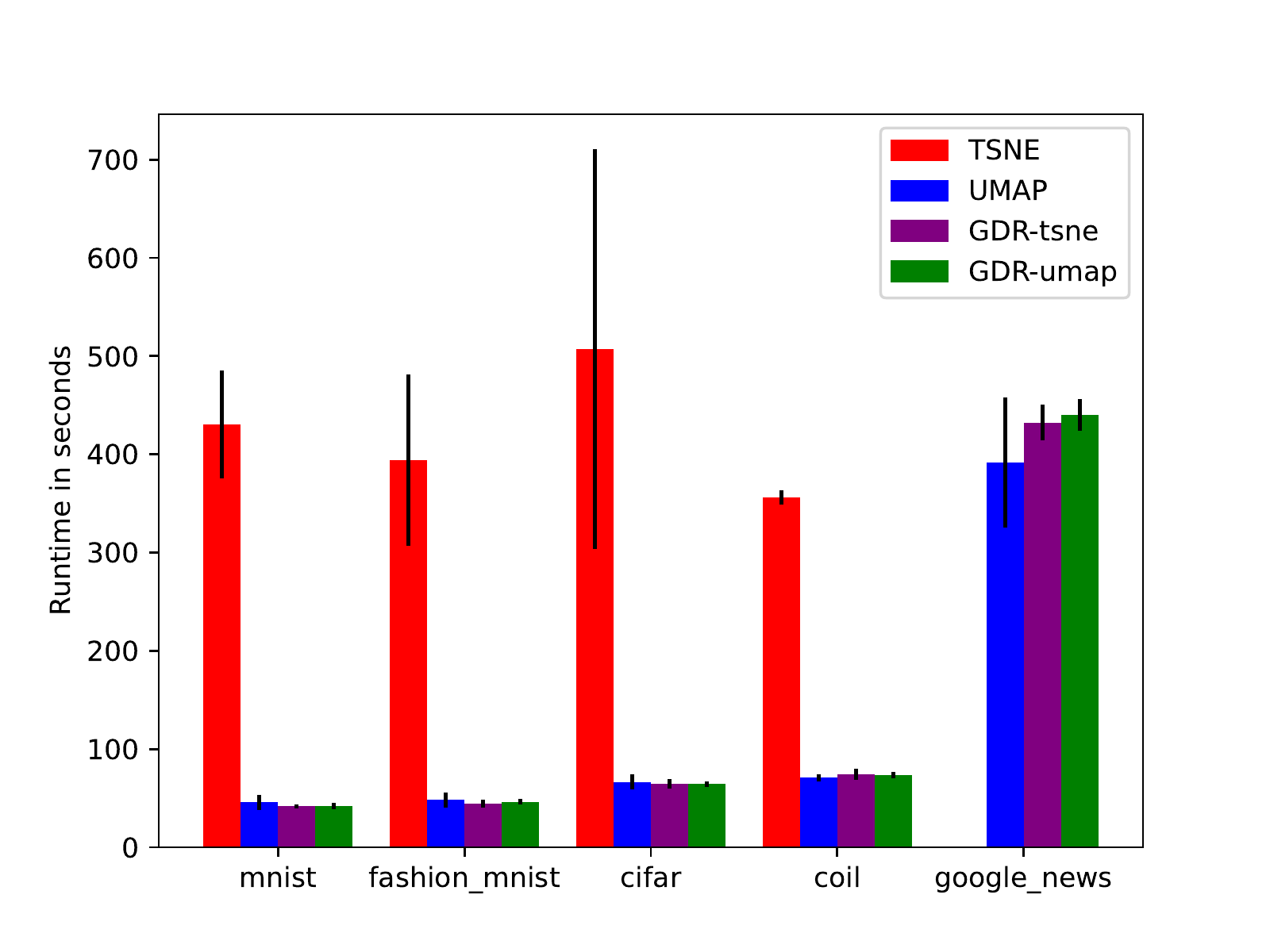}
    \caption{CPU runtimes for TSNE, UMAP, \ourmethodN, and \ourmethodU on common benchmark datasets. Note that we did not run TSNE on Google News as it does not converge in a reasonable amount of time.}
    \label{cpu_runtimes}
\end{figure}
\begin{figure}[!htb]
    \includegraphics[width=.95\linewidth]{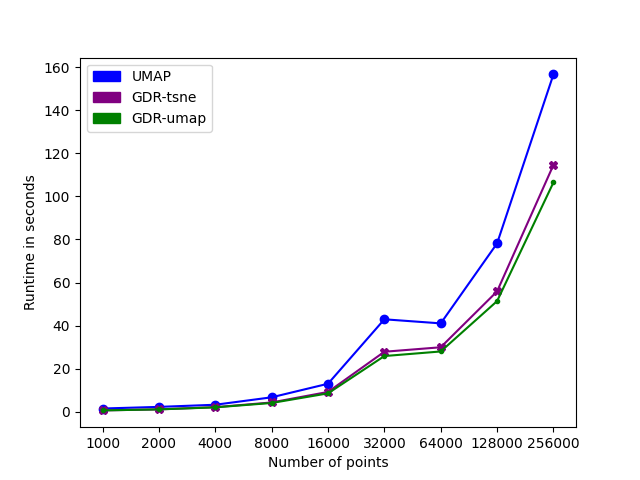}
    \caption{Runtimes for our implementations of TSNE and UMAP vs. \ourmethodN, and \ourmethodU as the size of MNIST grows. Up/Down sampling was done uniformly over points and noise was added when upsampling so as to avoid duplicates. All methods ran for 500 epochs. We did not show TSNE runtimes here as they do not converge on the larger sizes.}
    \label{data_size_runtimes}
\end{figure}

\begin{figure}[!htb]
    \includegraphics[width=.95\linewidth]{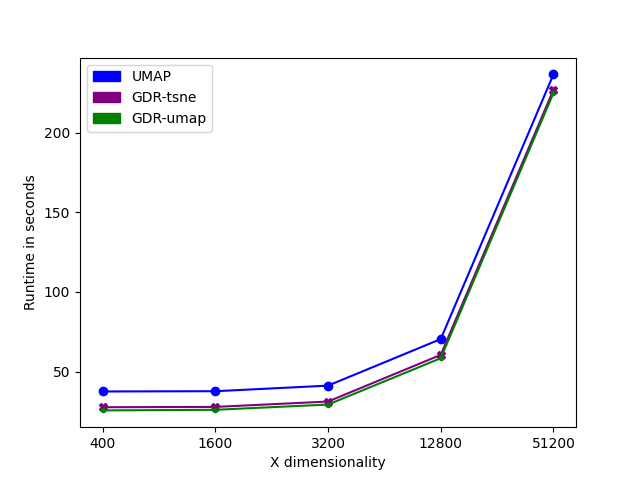}
    \caption{Runtimes for our implementations of TSNE and UMAP vs. \ourmethodN, and \ourmethodU as the dimensionality of MNIST grows. Up/Down sampling was done uniformly at random along the 784 dimensions. All methods ran for 500 epochs.}
    \label{dimensionality_runtimes}
\end{figure}

\subsection{Runtime Analysis}
We now turn to analyzing our method's runtime vs. that of TSNE and UMAP. Our implementations of TSNE and UMAP compute the exact same pre-processing steps and perform the same approximate nearest neighbor calculations. Therefore, we stick with the precedent set in~\cite{mcinnes2018umap} and~\cite{tang2016visualizing} and simply compare total runtimes rather than accounting for the various nearest-neighbor search implementations.
As a side note, our implementations of UMAP and TSNE are about $2\times$ faster than the standard implementations. We compare our code with the originals in Table~\ref{original_mnist_speeds}.

We see that \ourmethod consistently reaches state-of-the-art runtimes as compared to the existing dimensionality reduction algorithms. \ourmethod operates at least as fast as our optimized UMAP implementation and several orders of magnitude faster than our optimized Barnes-Hut TSNE implementation. This is due to multiple improvements over the existing algorithms. First, we have removed TSNE's reliance on the Barnes-Hut trees to calculate repulsions across all of the points, instead only calculating $\bigO(1)$ repulsions and scaling the resulting force appropriately. This resolves the primary speed discrepancy between the TSNE and UMAP optimization loops. Second, the symmetric attraction operation requires additional force computations -- deprecating this choice eases gradient calculations and memory access. Additionally, \ourmethod sets the scalars $a=b=1$, significantly simplifying gradient computations, especially under the Frobenius norm loss function. Lastly, by uniformly calculating one repulsion for each attraction, we remove UMAP's need for sampling multiple repulsions for each point while also simplifying the load distribution when parallelizing.

Figure~\ref{data_size_runtimes} shows that \ourmethod is consistent across data sizes. Furthermore, our speed over UMAP visibly improves as the dataset grows. The same is not the case for the dimensionality of the dataset, as evidenced in Figure~\ref{dimensionality_runtimes}. This is due to the fact that our speed improvement occurs during the optimization process, which is largely decoupled from the dimensionality of the high-dimensional dataset. We also mention that UMAP is significantly faster than any available TSNE implementation. Thus, performing as well as UMAP is a sufficient condition for obtaining the fastest available implementation of TSNE. We refer the reader to~\cite{mcinnes2018umap} for further analysis of the speed comparisons between UMAP and various TSNE methods.

We remark that UMAP obtains speed improvements due to the scalar sampling procedure, wherein they only sample attractions and repulsions according to $p_{ij}$ and $1 - p_{ik}$ (the latter of which is only ever estimated). \ourmethod, however, operates at the same speed as UMAP despite explicitly calculating attractions and repulsions.

\section{Conclusion}

We presented \ourmethod, a dimensionality reduction algorithm that can recreate TSNE and UMAP by toggling just one hyperparameter. This came as a result of studying all of the algorithmic differences between TSNE and UMAP and identifying which ones are responsible for the speed and embedding discrepancies. By analyzing each of these hyperparameters, we obtain speed at least as fast as UMAP but with greater flexibility than either previous method.

During this analysis, we also raised several questions regarding the theoretical implications that this hyperparameter analysis has on gradient-descent dimensionality reduction algorithms. Namely, we believe that many of the identified choices can be revisited. For example, does the KL divergence induce a better optimization criterium than the Frobenius norm? What other loss functions should one investigate? Are there other similarity measures $p$ and $q$ that permit simpler loss functions or optimization procedures? We believe that through identifying all of the components of the gradient-descent algorithms, we facilitated future research into identifying optimal choices for each.

\balance
\bibliographystyle{ACM-Reference-Format}
\bibliography{main}

\end{document}